\documentclass[10pt]{article}

\usepackage{scai}

\usepackage[authoryear, sort&compress, round]{natbib}
\usepackage{xspace}
\usepackage{adjustbox}
\usepackage{array}
\usepackage{float}
\usepackage{dblfloatfix}
\usepackage{setspace}
\usepackage{footmisc}
\usepackage{booktabs}
\usepackage{multirow}
\usepackage{longtable}
\usepackage{tabularx}
\usepackage{xltabular}
\usepackage{threeparttable}
\usepackage{siunitx}
\usepackage{mathtools}
\usepackage{bm}
\usepackage{dsfont}
\usepackage{subcaption}
\usepackage{tikz}
\usepackage{pgfplots}
\pgfplotsset{compat=1.18}
\usepackage{algorithm}
\usepackage{algpseudocode}
\usepackage{listings}
\usepackage{cleveref}
\usepackage{todonotes}
\usepackage{fontawesome5}
\usepackage{CJKutf8}
\usepackage{lipsum}
\usepackage{arydshln}
\usepackage{multicol}
\usepackage{titletoc}
\usepackage{mdframed}
\usepackage{tabularray}
\UseTblrLibrary{booktabs}
\usepackage[table,dvipsnames]{xcolor}
\usepackage{makecell}
\usepackage{dashbox}

\newcolumntype{L}[1]{>{\raggedright\let\newline\\\arraybackslash\hspace{0pt}}m{#1}}
\newcolumntype{R}[1]{>{\raggedleft\let\newline\\\arraybackslash\hspace{0pt}}m{#1}}

\newcommand{\ignore}[1]{}

\makeatletter
\DeclareRobustCommand\onedot{\futurelet\@let@token\@onedot}
\def\@onedot{\ifx\@let@token.\else.\null\fi\xspace}

\makeatother

\definecolor{MyBlue}{rgb}{0.46, 0.50, 0.61}
\definecolor{MyDarkBlue}{rgb}{0,0.08,0.8}
\definecolor{MyDarkGreen}{RGB}{45,155,45}
\definecolor{MyDarkRed}{rgb}{0.8,0.02,0.02}
\definecolor{MyOrange}{rgb}{1.0, 0.4, 0.2}
\definecolor{MyPurple}{RGB}{111,0,255}
\definecolor{MyRed}{rgb}{0.8,0.0,0.0}
\definecolor{MyGold}{rgb}{0.75,0.6,0.12}
\definecolor{MyDarkgray}{rgb}{0.66, 0.66, 0.66}
\definecolor{MyBrown}{rgb}{0.65, 0.16, 0.16}
\definecolor{MyMutedRose}{rgb}{0.58, 0.29, 0.35}
\definecolor{JiayuanColor}{rgb}{0.60,0.43,0.48}
\definecolor{erranColor}{rgb}{24, 40, 113}

\definecolor{citecolor}{HTML}{696FAD}

\newif\ifpropositionfirstitem
\propositionfirstitemtrue

\definecolor{bggray}{HTML}{F5F5F5}
\definecolor{pvdblue}{HTML}{DAE8FC}
\definecolor{RoseQuartzBg}{HTML}{F7CAC9}
\definecolor{RoseQuartz}{HTML}{F5A798}
\definecolor{Serenity}{HTML}{92A8D1}
\definecolor{OrangeRed}{rgb}{1.0, 0.27, 0.0}
\definecolor{RoyalBlue}{cmyk}{1, 0.50, 0, 0}
\definecolor{Turquoise}{HTML}{0F4C81}
\definecolor{mint}{rgb}{0.24, 0.71, 0.54}
\definecolor{green}{rgb}{0.0, 0.120, 0.0}

\newdimen\abovecrulesep
\newdimen\belowcrulesep
\makeatletter
\patchcmd{\@@@cmidrule}{\aboverulesep}{\abovecrulesep}{}{}
\patchcmd{\@xcmidrule}{\belowrulesep}{\belowcrulesep}{}{}
\makeatother

\definecolor{mybluetitle}{HTML}{4B527E} 

\definecolor{codegreen}{HTML}{478058}
\definecolor{codegray}{rgb}{0.5,0.5,0.5}
\definecolor{codepurple}{HTML}{4F5E80} 
\definecolor{backcolour}{rgb}{0.95,0.95,0.95}
\lstdefinestyle{mystyle}{
    backgroundcolor=\color{backcolour},
    commentstyle=\color{codegreen},
    keywordstyle=\color{magenta},
    numberstyle=\tiny\color{codegray},
    stringstyle=\color{codepurple},
    basicstyle=\ttfamily\scriptsize,
    breakatwhitespace=false,
    breaklines=true,
    captionpos=b,
    keepspaces=true,
    frame=none,
    numbersep=5pt,
    showspaces=false,
    showstringspaces=false,
    showtabs=false,
    tabsize=2
}

\newtcolorbox{promptbox}[2][]{
    enhanced, 
    breakable,
    center title,
    left*=0pt, right*=0pt,
    boxsep=2pt, left=5pt, right=5pt,
    skin first=enhanced,
    skin middle=enhanced,
    skin last=enhanced,
    colback  = backcolour,
    fonttitle=\bfseries\rmfamily,
    fontupper=\scriptsize,
    title={\footnotesize\strut{#2}},
    #1
    }

\newtcolorbox{onebox}[2][]{
    enhanced, 
    center title,
    left*=0pt, right*=0pt,
    boxsep=2pt, left=5pt, right=5pt,
    skin first=enhanced,
    skin middle=enhanced,
    skin last=enhanced,
    colframe = mybluetitle!90,
  colback  = mybluetitle!10,
    fonttitle=\bfseries\rmfamily\fontfamily{phv}\selectfont,
    title={\strut{#2}  \refstepcounter{subsubsection} \addcontentsline{toc}{subsubsection}{\string\numberline{\thesubsubsection}#2}
    },
    #1
    }

\definecolor{violet}{RGB}{111,45,168}
\definecolor{rqblue}{RGB}{45,80,130}

\definecolor{MyGreen}{RGB}{0,128,0}
\definecolor{darkgrey}{rgb}{0.25, 0.25, 0.25}

\definecolor{uclablue}{HTML}{2774AE}
\definecolor{uclagold}{HTML}{FFD100}
\definecolor{uclanavy}{HTML}{003B5C}
\definecolor{uclabluebg}{HTML}{EEF5FA}
\definecolor{uclagoldbg}{HTML}{FFF9E0}
\definecolor{uclagolddk}{HTML}{8A6D00}   

\newcommand{\researchq}[3]{%
  \par\smallskip
  \noindent
  \textbf{\color{rqblue}RQ#1}
  \hspace{0.4em}
  \textbf{(#2)}
  \hspace{0.4em}
  #3
  \par\smallskip
}

\newtcolorbox[auto counter]{finding}[2][]{%
  enhanced, breakable,
  colback=uclabluebg, colframe=uclablue,
  boxrule=0.6pt, arc=3pt,
  left=7pt, right=7pt, top=9pt, bottom=5pt,
  fonttitle=\bfseries\small, fontupper=\small,
  coltitle=white, colbacktitle=uclablue,
  attach boxed title to top left={xshift=8pt, yshift=-\tcboxedtitleheight/2},
  boxed title style={colframe=uclablue, arc=2pt, boxrule=0pt,
                     left=5pt, right=5pt, top=1.5pt, bottom=1.5pt},
  title={Finding~\thetcbcounter\;\textbar\;#2}, #1}

\definecolor{violet}{RGB}{111,45,168}
\annotator{mandy}{brown}
\annotator{haixin}{olive}
\annotator{wei}{blue}
\annotator{YS}{red}

\title{ACLArena: Agent Continual Learning in Multi-stage Post-training}

\scaiauthor{1,lead}{Haixin Wang}
\scaiauthor{1,lead}{Xiaoxuan Wang}
\scaiauthor{1}{Junkai Zhang}
\scaiauthor{1}{Han Zhang}
\scaiauthor{1}{Renliang Sun}
\scaiauthor{1}{Alexander Taylor}
\scaiauthor{1}{Yidan Shi}
\scaiauthor{1}{Haoran Deng}
\scaiauthor{2}{Chenguang Wang}
\scaiauthor{1}{Jason Cong}
\scaiauthor{1}{Yizhou Sun}
\scaiauthor{1}{Wei Wang}

\affiliation{1}{University of California, Los Angeles}
\affiliation{2}{University of California, Santa Cruz}
\contribnote{lead}{Equal Contribution.}

\metadata[GitHub]{\url{https://github.com/WillDreamer/ACLArena}}
\metadata[HuggingFace]{\url{https://huggingface.co/collections/willhx/aclarena}}
\correspondence{Xiaoxuan Wang}{xw27@g.ucla.edu}

\begin{abstract}
Building general-purpose agents for industrial deployment requires integrating multiple capabilities, each typically acquired at a distinct stage of training. 
Yet there is currently no well-established recipe for \textbf{Agent Continual Learning (ACL)}, with little understanding of the trade-offs among existing integration paradigms. To address this gap, we introduce ACLArena, a framework for comprehensively studying, analyzing, and evaluating ACL. We first build a sequential training pipeline and conduct an in-depth analysis that explains the mechanisms of forgetting and generalization from two complementary perspectives, the model level and the token level. 
Guided by these analyses, we systematically compare multi-teacher on-policy distillation, self-distilled fine-tuning, and model merging to assess their ability to recover previously learned capabilities while preserving newly acquired ones.
Through extensive experiments, we develop a detailed understanding of how capabilities transfer across stages.
Finally, we propose a new ACL recipe that combines offline replay over high-quality trajectories with a routed network of multiple LoRA experts each specialized via RL, substantially improving the agent's ability to learn across multiple domains. Comprehensive experiments on four reasoning and agentic tasks, evaluated under both in-domain and out-of-domain settings, demonstrate the value of our analysis and the effectiveness of our approach.
\end{abstract}

\begin{document}

\maketitle

\begin{figure}[!htbp]
    \centering
    \vskip-0.7em
    \includegraphics[width=.99\linewidth]{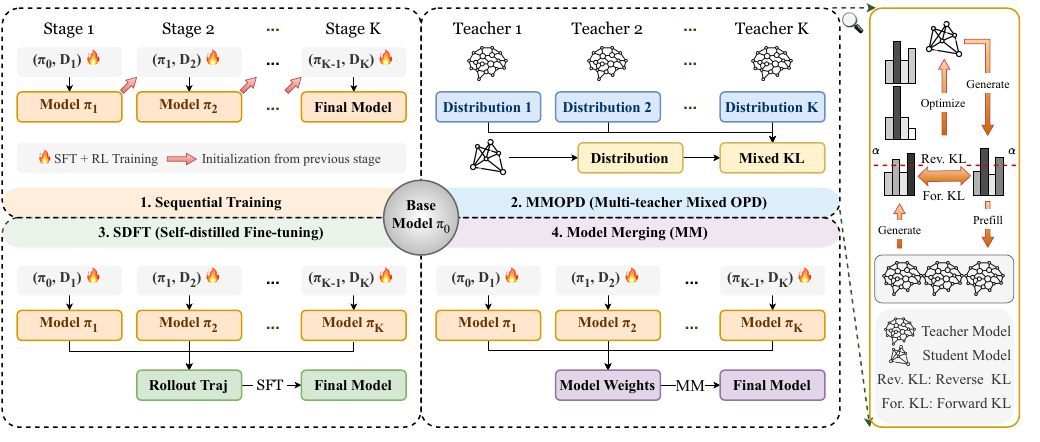}
    \caption{Overview of the four mainstream paradigms for capability consolidation: sequential training, multi-teacher mixed on-policy distillation (MMOPD), self-distilled fine-tuning (SDFT), and model merging (MM).}
    \label{fig:main}
    \vskip-7em
\end{figure}
\clearpage

\section{Introduction}

Modern agentic LLMs rely on multi-stage post-training, where capabilities such as CoT reasoning, tool use, and instruction following are acquired sequentially through heterogeneous training pipelines and environments~\citep{yang2025qwen3,zeng2026glm,deepseekai2026deepseekv4highlyefficientmilliontoken}. That is, training an industrial agent is fundamentally a continual learning problem: the model must develop new capabilities from new environments while retaining those acquired before. This is precisely why recent technical reports have increasingly explored diverse training strategies to expand the capabilities of general-purpose agents across multiple domains. 

However, there are still some limitations in this area. First, these reports provide limited training details and rarely compare alternatives under controlled settings. Second, many classical continual-learning methods are impractical for agent training: replay requires retaining costly environments and reward pipelines~\citep{rebuffi2017icarl,lopez2017gradient}, task-specific modules complicate deployment, and joint retraining becomes increasingly expensive as capabilities accumulate. Consequently, ACL still lacks a realistic testbed and a controlled comparison of practical training strategies.

To advance the study of ACL, we develop a pipeline to systematically investigate what happens when an LLM agent sequentially undergoes multiple post-training stages in substantially different environments, uncover the underlying causes of the observed phenomena, and explore how they can be effectively addressed. Our investigation is organized around three progressively deeper research questions:
\researchq{1}{Phenomenon}{
How does sequential post-training reshape an agent's capabilities? To what extent does learning new capabilities cause forgetting of previously acquired ones, and when does it instead enable generalization to out-of-domain tasks?
}
\researchq{2}{Mechanism}{
What fundamentally drives forgetting and transfer across post-training stages? How do changes in the agent's behavior, representations, and optimization interact to determine whether capabilities interfere with or reinforce one another?
}
\researchq{3}{Method}{
Can prior LLMs be effectively reused to mitigate forgetting and promote transfer, either in parameter space or in behavior space through the trajectories they generate?
}

To address these questions, our \textsc{ACLArena} first starts from a base model and progresses through multiple stages of post-training with a \uline{sequential training} pipeline as the baseline for RQ1. We use the simplest route and the default in many practical pipelines as a microscope on the phenomenon itself. Building on this setup, we conduct an in-depth analysis from two complementary perspectives: model-wise, by tracking how capabilities evolve across stages, and token-wise, by examining how individual training signals shape these dynamics for RQ2.

Building on the diagnoses, we summarize the existing mainstream technical approaches and unify them under our framework for RQ3. Specifically, we study three principled strategies that reuse prior models along different dimensions: \uline{on-policy distillation (OPD)} reuses prior models in behavior space online, letting teacher checkpoints supervise the student on its own state distribution; \uline{self-distilled fine-tuning (SDFT)} reuses them in behavior space offline, replaying filtered specialist trajectories as data; and \uline{model merging (MM)} reuses them directly in parameter space. For a fair comparison, we tune each route to its best attainable configuration rather than evaluating strawman implementations.

Our investigation leads to three findings that guide the design of our method. First, task-specific post-training induces distinct and only partially aligned optimization directions, resulting in uneven transfer and forgetting across tasks. Second, consolidating heterogeneous capabilities within a single shared model creates persistent trade-offs, as recovering one capability can degrade others. Third, SFT and RL play complementary roles: SFT establishes valid task behavior through broader and more directionally consistent updates, whereas RL makes smaller, policy-local updates that refine task competence. Together, these findings point to the need for an ACL paradigm that reconciles task-specific adaptation with capability consolidation while reducing cross-task interference.


Based on these findings, we further propose a novel ACL paradigm, \uline{Mixture of Low-Rank Experts (MLE)}, which, after rapidly learning the behaviors of multiple domains via SDFT, leverages LoRA adapters to perform RL for the final capability refinement. Each stage thus yields its own set of experts, and a routing mechanism composes them so that a single model can serve the tasks of all stages. This paradigm serves multi-stage post-training within a single model and substantially mitigates forgetting.

In summary, our contributions are threefold:
\begin{itemize}[leftmargin=*]
\item \textbf{A realistic benchmark.} We construct \textsc{ACLArena}, a multi-stage post-training pipeline with heterogeneous environments to reveal forgetting and generalization.
\item \textbf{A mechanistic study.} Using sequential training as a diagnostic baseline, we characterize forgetting and transfer, uncover their underlying mechanisms, and provide optimized comparisons of three practical ACL paradigms.
\item \textbf{A practical mitigation strategy.} Guided by our analysis, we combine their respective strengths into a unified ACL method that performs competitively across heterogeneous tasks and consistently improves capability integration
\end{itemize}

\section{Related Work}

\subsection{Multi-teacher OPD}
Recent LLM post-training pipelines increasingly adopt OPD as a unifying stage for consolidating improvements obtained from multiple stages. OPD implements this by converting the student’s own sampled trajectories into dense token-level supervision under stronger teacher policies~\citep{agarwal2024onpolicydistillationlanguagemodels,gu2026minillmonpolicydistillationlarge,lu2025onpolicydistillation}. 
Several recent models exemplify this trend. Qwen3~\citep{yang2025qwen3} performs OPD by first sampling prompts and letting the student generate responses, then minimizing the KL divergence between the student and teacher distributions on these self-generated trajectories. MiMo-V2~\citep{xiao2026mimo} extends this paradigm to a multi-teacher setting, formulating Multi-Teacher OPD as an on-policy reinforcement learning objective. GLM-5~\citep{zeng2026glm} employs OPD as a final refinement stage to mitigate capability regression while preserving gains accumulated across earlier training phases. Similarly, Nemotron-Cascade 2~\citep{yang2026nemotron} introduces multi-domain OPD by selecting the strongest validation checkpoint from each Cascade RL benchmark category as a capability-diverse teacher pool. DeepSeek-V4~\citep{deepseekai2026deepseekv4highlyefficientmilliontoken} further uses multi-teacher OPD as the primary mechanism for merging expert capabilities into the final model, and adopts full-vocabulary logit distillation to reduce the variance.


\subsection{Self-distilled Fine-tuning}

SDFT consolidates specialist capabilities through filtered teacher-generated traces and a subsequent alignment-oriented RL stage, offering a simpler and more stable training recipe but relying more heavily on data-mixture design and providing weaker on-policy correction than OPD. MAI-Thinking-1~\citep{microsoftai2026maithinking1} performs SDFT by rejection-sampling and lightly filtering rollouts from multiple checkpoints of specialist teacher models, then apply a lightweight RL stage focused on safety, over-refusal reduction, and style while retaining some STEM/coding data to preserve reasoning performance.

\subsection{Model Merging}
A recent public model release, Rio 3.5 Open 397B~\citep{rio2026}, provides an illustrative example of a merge-then-distill pipeline for LLMs. According to the model documentation, the released system was constructed by merging Nex-N2-Pro~\citep{nexn2pro} with Qwen3.5-397B-A17B~\citep{qwen3.5}, followed by OPD from a stronger teacher model.It is useful as a real-world instance of combining weight-space model merging with post-hoc distillation to consolidate capabilities into a single deployed model.

\begin{table*}[t]
\centering
\caption{
Main sequential training results.
Single-hop Search reports the average performance on PopQA and TriviaQA.
Multi-hop Search reports the average performance on 2Wiki, Bamboogle,
HotpotQA, and MuSiQue. Dashed boxes indicate the training stage corresponding to each task, while changes in the subsequent rows reflect generalization and forgetting. We denote the checkpoints after each training stage as \textsc{Seq-Math}, \textsc{Seq-Search}, \textsc{Seq-E-commerce}, and \textsc{Seq-IF}, respectively; \textsc{Seq-IF} is the final sequential checkpoint and is referred to as \textsc{Seq-Final} in subsequent experiments. The best score in each column is highlighted in \textbf{\textcolor{red}{red}}. All results are reported as mean $\pm$ standard deviation over three independent runs.
}
\label{tab:mainseq}
\setlength{\tabcolsep}{3.5pt}
\renewcommand{\arraystretch}{1.05}
\resizebox{\textwidth}{!}{%
\begin{tabular}{l cccc ccccccc}
\toprule
& \multicolumn{4}{c}{\textbf{In-domain evaluation} ($\uparrow$)}
& \multicolumn{7}{c}{\textbf{Out-of-domain evaluation} ($\uparrow$)} \\
\cmidrule(lr){2-5}
\cmidrule(lr){6-12}

& \multirow{2}{*}{\shortstack{AIME26\\(avg@16)}}
& \multirow{2}{*}{NQ}
& \multirow{2}{*}{$\tau^3$-Retail}
& \multirow{2}{*}{IF-Eval}
& \multicolumn{2}{c}{Math}
& \multicolumn{1}{c}{Single-hop Search}
& \multicolumn{1}{c}{Multi-hop Search}
& \multicolumn{2}{c}{E-commerce}
& \multicolumn{1}{c}{IF} \\

\cmidrule(lr){6-7}
\cmidrule(lr){8-8}
\cmidrule(lr){9-9}
\cmidrule(lr){10-11}
\cmidrule(lr){12-12}

\textbf{Pipeline}
& & & &
& GPQA & MMLU
& Avg.
& Avg.
& $\tau^3$-Telecom & $\tau^3$-Mock
& IF-Bench \\

\midrule

\rowcolor{uclablue}
\multicolumn{12}{l}{
    \textcolor{white}{\emph{Per-stage performance} 
    (Math $\rightarrow$ Search $\rightarrow$ E-commerce $\rightarrow$ Instruction Following)}
} \\

\midrule

Base
& $6.46$
& $13.3_{\scriptscriptstyle \pm 0.2}$
& $3.5_{\scriptscriptstyle \pm 0.4}$
& $46.0_{\scriptscriptstyle \pm 0.9}$
& $30.8_{\scriptscriptstyle \pm 1.6}$
& $65.1_{\scriptscriptstyle \pm 0.3}$
& $22.3_{\scriptscriptstyle \pm 0.0}$
& $11.1_{\scriptscriptstyle \pm 0.6}$
& $17.5_{\scriptscriptstyle \pm 1.0}$
& $5.0_{\scriptscriptstyle \pm 2.5}$
& $14.4_{\scriptscriptstyle \pm 0.2}$ \\

\midrule

Seq-${\text{Math}}$
& \dashbox{$\textcolor{red}{\mathbf{25.83}}$}
& $22.0_{\scriptscriptstyle \pm 0.2}$
& $2.3_{\scriptscriptstyle \pm 0.5}$
& $43.3_{\scriptscriptstyle \pm 0.6}$
& \dashbox{$\textcolor{red}{\mathbf{40.2_{\scriptscriptstyle \pm 2.5}}}$}
& \dashbox{$79.0_{\scriptscriptstyle \pm 0.2}$}
& $38.7_{\scriptscriptstyle \pm 0.2}$
& $22.8_{\scriptscriptstyle \pm 0.5}$
& $13.1_{\scriptscriptstyle \pm 2.1}$
& $3.3_{\scriptscriptstyle \pm 3.8}$
& $17.1_{\scriptscriptstyle \pm 0.2}$ \\

Seq-${\text{Search}}$
& $23.33$
& \dashbox{$\textcolor{red}{\mathbf{45.2_{\scriptscriptstyle \pm 0.4}}}$}
& $2.0_{\scriptscriptstyle \pm 0.3}$
& $39.8_{\scriptscriptstyle \pm 0.2}$
& $37.2_{\scriptscriptstyle \pm 0.7}$
& $79.2_{\scriptscriptstyle \pm 0.1}$
& \dashbox{$\textcolor{red}{\mathbf{52.9_{\scriptscriptstyle \pm 0.1}}}$}
& \dashbox{$\textcolor{red}{\mathbf{37.4_{\scriptscriptstyle \pm 0.8}}}$}
& $9.1_{\scriptscriptstyle \pm 3.9}$
& $30.8_{\scriptscriptstyle \pm 1.4}$
& $15.8_{\scriptscriptstyle \pm 0.2}$ \\

Seq-${\text{E-commerce}}$
& $6.04$
& $14.6_{\scriptscriptstyle \pm 0.1}$
& \dashbox{$\textcolor{red}{\mathbf{34.0_{\scriptscriptstyle \pm 0.9}}}$}
& $41.3_{\scriptscriptstyle \pm 0.3}$
& $36.0_{\scriptscriptstyle \pm 1.4}$
& $67.3_{\scriptscriptstyle \pm 0.3}$
& $13.2_{\scriptscriptstyle \pm 0.1}$
& $9.4_{\scriptscriptstyle \pm 0.2}$
& \dashbox{$\textcolor{red}{\mathbf{46.3_{\scriptscriptstyle \pm 1.6}}}$}
& \dashbox{$\textcolor{red}{\mathbf{74.2_{\scriptscriptstyle \pm 1.4}}}$}
& $19.3_{\scriptscriptstyle \pm 0.9}$ \\

Seq-${\text{IF}}$ (=Seq-${\text{Final}}$)
& $10.21$
& $33.5_{\scriptscriptstyle \pm 1.7}$
& $29.6_{\scriptscriptstyle \pm 1.6}$
& \dashbox{$\textcolor{red}{\mathbf{84.8_{\scriptscriptstyle \pm 0.1}}}$}
& $37.9_{\scriptscriptstyle \pm 1.7}$
& $\textcolor{red}{\mathbf{79.5_{\scriptscriptstyle \pm 0.0}}}$
& $44.9_{\scriptscriptstyle \pm 0.7}$
& $25.0_{\scriptscriptstyle \pm 1.5}$
& $45.9_{\scriptscriptstyle \pm 1.8}$
& $62.5_{\scriptscriptstyle \pm 6.6}$
& \dashbox{$\textcolor{red}{\mathbf{30.8_{\scriptscriptstyle \pm 0.2}}}$} \\

\bottomrule
\end{tabular}%
\vspace{-0.3cm}
}
\end{table*}

\section{Diagnosing Baseline: Sequential Training}


\subsection{Multi-stage Task Curation}
We construct a representative multi-stage post-training pipeline. We deliberately study a fixed curriculum rather than arbitrary task permutations. 
Our goal is to model a realistic multi-stage post-training pipeline, where stages are typically introduced according to capability prerequisites rather than in a randomly permuted order. 
Specifically, we adopt mathematical reasoning as the first-stage task to strengthen the model’s reasoning capability. The second stage focuses on agentic tasks, and the model first learns search, which involves a single tool before progressing to E-commerce, which requires up to fifteen tools. 
Finally, Instruction Following (IF) is introduced as the last stage to improve instruction adherence and alignment. (Math$\rightarrow$Search$\rightarrow$E-commerce$\rightarrow$IF)

Since the pretrained base model already possesses basic reasoning ability, both the math and IF stages are trained by RL directly using critic-free algorithm~\citep{shao2024deepseekmath,wang2026arlarena}. Since agentic tasks require structured tool invocation that is absent in the base model, we first perform rejection-sampled fine-tuning as a cold-start stage to teach the model the required tool-use format and interaction protocol, followed by critic-free RL to further improve decision-making and long-horizon interaction performance.

\subsection{Sequential Training Formulation}

Suppose the post-training pipeline consists of $K$ sequential stages. Starting from a pretrained base policy $\pi_0$, the model is optimized progressively in a curriculum manner, with each stage introducing increasingly complex capabilities. Upon completing all $K$ stages, the resulting checkpoint undergoes a final safety alignment stage before deployment. At stage $k \in \{1,\ldots,K\}$, the current policy $\pi_{k-1}$ is trained on inputs drawn from a stage-specific prompt distribution $\mathcal{D}_k$ and is optimized with respect to a stage-specific reward function $r_k$, yielding an updated policy $\pi_k$. For agentic stages, the policy additionally interacts with a stage-specific environment $\mathcal{E}_k$, which returns tool observations and simulated-user turns in response to its actions; throughout the paper, $\mathcal{D}_k$ always denotes this input distribution alone, never the environment or a materialized dataset.
The overall sequential training process is illustrated as follows:
\[
\pi_0
\overset{\mathcal{D}_1}{\longrightarrow}
\pi_1
\overset{\mathcal{D}_2}{\longrightarrow}
\cdots
\overset{\mathcal{D}_K}{\longrightarrow}
\pi_K,
\]
where $\pi_0$ denotes the pretrained base model and $\pi_k$ represents the checkpoint obtained after completing the $k$-th post-training stage, which is carried out on $\mathcal{D}_k$, paired with the environment $\mathcal{E}_k$ for the agentic stages.

\begin{figure*}[t]
\centering
\begin{subfigure}[t]{0.24\linewidth}
\centering
\includegraphics[width=\linewidth]{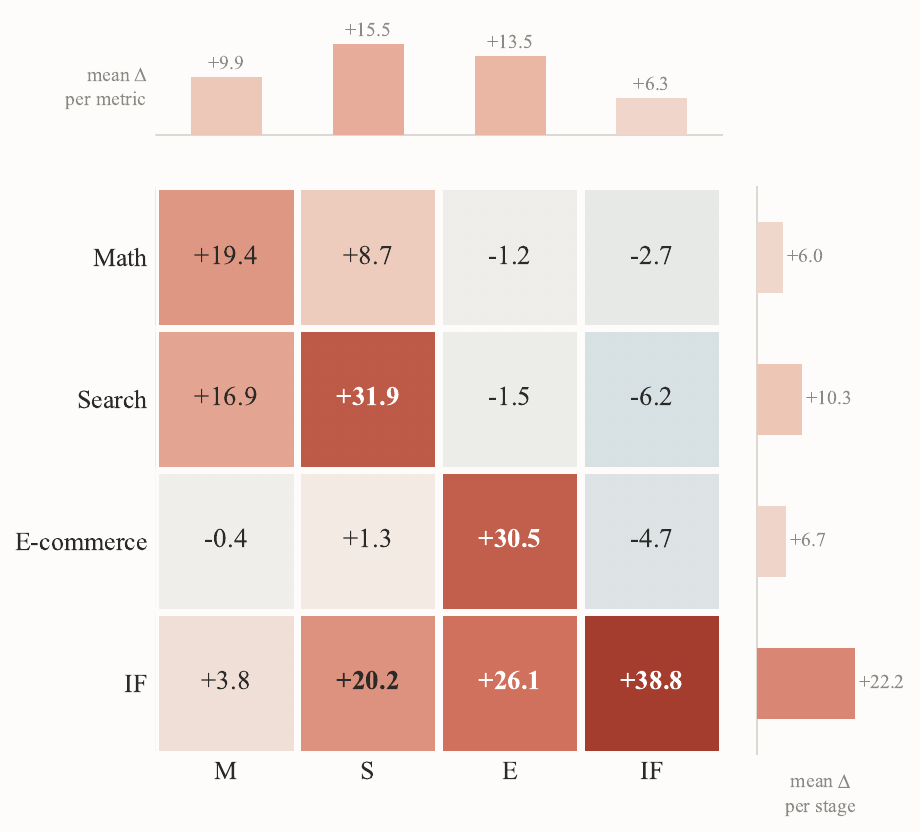}
\caption{In-domain evaluation change after sequential training (vs. base)}
\label{fig:a}
\end{subfigure}%
\hfill
\begin{subfigure}[t]{0.24\linewidth}
\centering
\includegraphics[width=\linewidth]{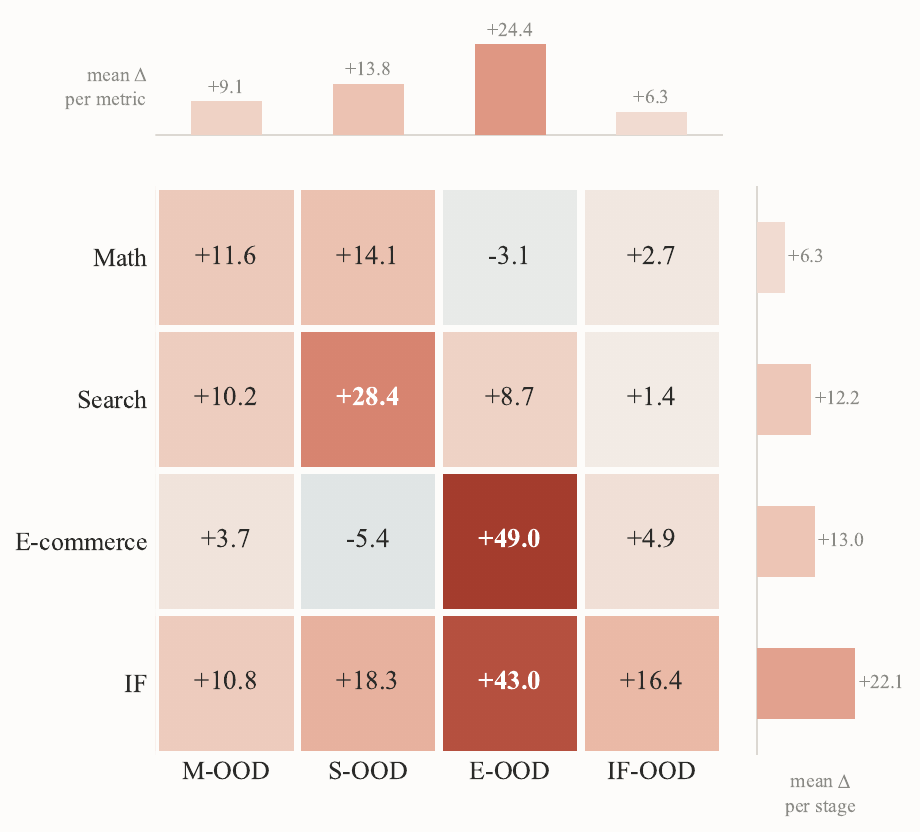}
\caption{Out-of-domain evaluation change after sequential training (vs. base)}
\label{fig:b}
\end{subfigure}%
\hfill
\begin{subfigure}[t]{0.24\linewidth}
\centering
\includegraphics[width=\linewidth]{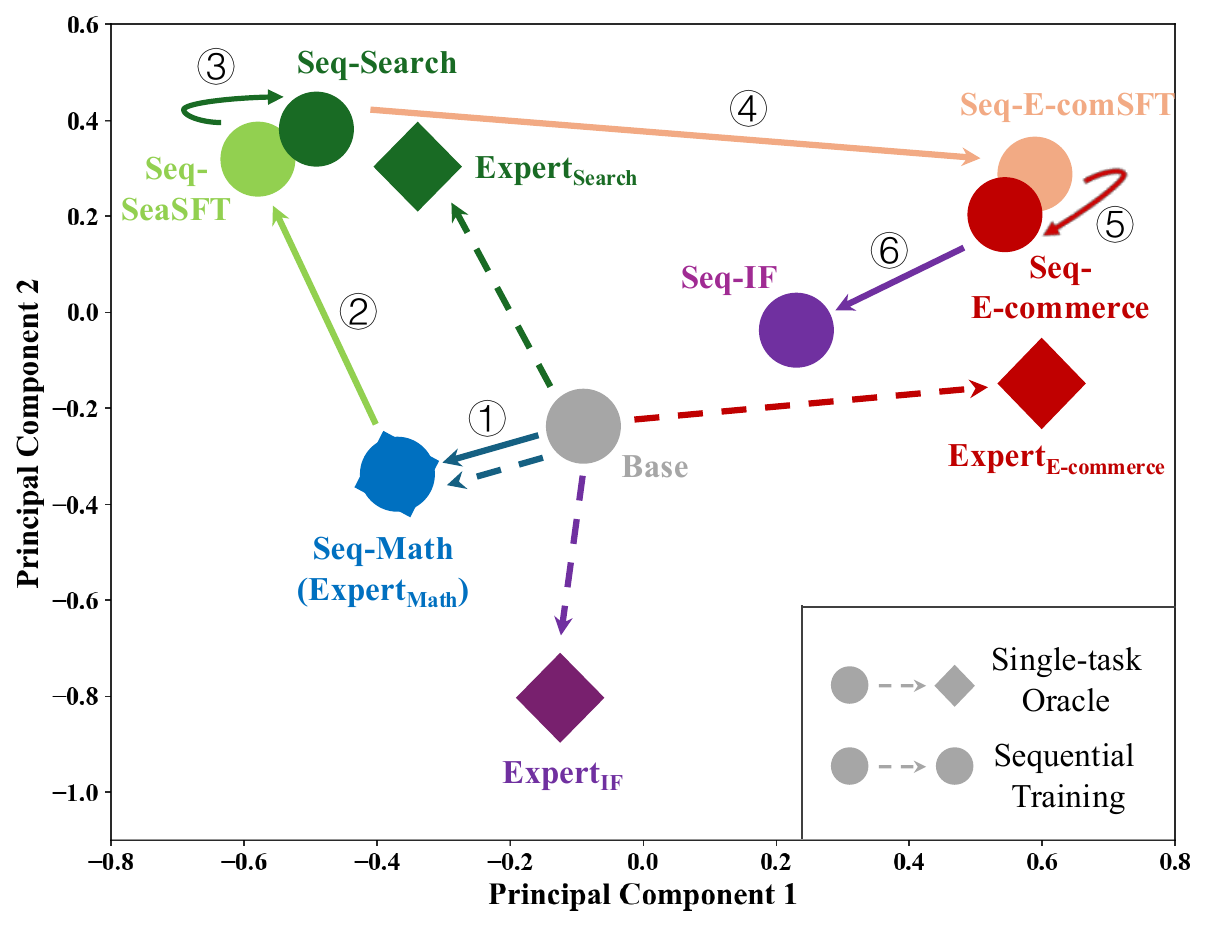}
\caption{Principal component analysis of model weights in different stages and the oracle ones.}
\label{fig:c}
\end{subfigure}%
\hfill
\begin{subfigure}[t]{0.24\linewidth}
\centering
\includegraphics[width=\linewidth]{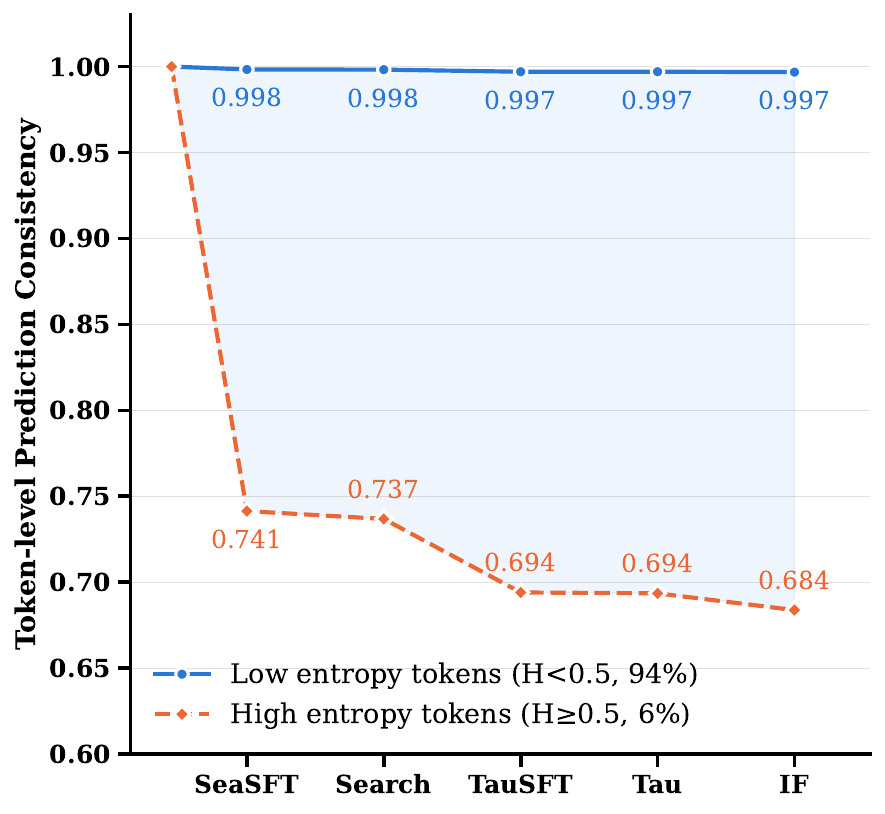}
\caption{The token distribution shift induced by the post-math stages (SFT and RL separated)}
\label{fig:d}
\end{subfigure}
\vspace{-0.2cm}
\caption{Diagnosing sequential training. Rows (top to bottom) follow the training order; columns correspond to the evaluation of each domain, and marginal bars show mean performance gap $\Delta$ per metric and per stage. The projection in (c) reveals both that the tasks occupy distinct optimization directions and that it progressively fails to recover them. (d) provides token-level analysis of original rollout with different entropy.}
\vspace{-0.3cm}
\label{fig:seq-ana}
\end{figure*}
\newpage
\subsection{Diagnosing Sequential Training}

\subsubsection{Main Results}

Table 1 reports how in-domain and out-of-domain performance evolves along the sequential training, revealing both positive transfer and substantial forgetting. Early stages exhibit beneficial cross-task transfer: Seq-${\text{Math}}$ improves NQ performance from 13.3 to 22.0 before search-specific training, while the subsequent Seq-${\text{Search}}$ largely preserves the acquired math capability. However, Seq-${\text{E-commerce}}$ introduces pronounced interference, reducing AIME and NQ scores from 23.33 to 6.04 and from 45.2 to 14.6, respectively. Similar regressions occur on out-of-domain search benchmarks, where multi-hop performance drops from 37.4 to 9.4. The final Seq-${\text{IF}}$ partially restores earlier capabilities, recovering NQ to 33.5 and multi-hop search to 25.0, while achieving 84.8 on IF-Eval. Overall, these results show that capability evolution under sequential training is non-monotonic and highly task-dependent: later stages can both disrupt and recover previously acquired behaviors.

\subsubsection{\textbf{Capability Dynamics: Forgetting and Transfer}} 

Figures~\ref{fig:a} and \ref{fig:b} provide a fine-grained view of how capabilities evolve as sequential training shifts across domains. \textit{Vertically}, the heatmaps track the dynamics of each capability across training stages. Every capability peaks at its own stage and degrades afterwards, but the severity of that decay is strongly task-dependent. Searching capability is acquired rapidly and then forgotten most abruptly: NQ falls from $45.2$ to $14.6$ as soon as E-commerce training begins, and the final stage restores it only to $33.5$. Math reasoning capability decays gradually and then collapses, from $25.83$ at its own stage to $23.33$ after search and $6.04$ after E-commerce, rebounding only to $10.21$. E-commerce capability is the most durable of the three, yet it too declines once training moves on, with $\tau^3$-Retail slipping from $34.0$ to $29.6$. No capability improves monotonically after the stage that instills it. The bar plots above the heatmaps further show that search achieves the largest immediate capability gain during its learning stage.

\textit{Horizontally}, the heatmaps characterize the model’s capability profile across domains after each training stage. As summarized by the bar plots on the right, this profile does not strengthen steadily as sequential training progresses. It broadens through the math and search stages, regresses sharply at the E-commerce stage, where AIME26, IF-Eval and both search splits fall back to or below the base model's level, and only partially recovers at the final IF stage. The resulting checkpoint does not dominate its predecessors either: AIME26 ($10.21$ vs. $25.83$), NQ ($33.5$ vs. $45.2$) and $\tau^3$-Retail ($29.6$ vs. $34.0$) all end below the peaks reached earlier in the curriculum. New capabilities are thus not simply accumulated on top of old ones; each stage partially overwrites what preceded it, highlighting substantial heterogeneity in capability retention across domains.

\subsubsection{\textbf{Model-level Diagnosis: Parameter Displacement.}} 
Figure \ref{fig:c} projects the unit-normalized checkpoints onto the first two principal components, so that a point's position reflects the direction of optimization.
The four single-task oracles exhibit clearly distinct parameter-displacement directions in the PCA projection, suggesting that different post-training tasks induce heterogeneous parameter updates: the update direction that best serves one task points away from those that serve the others, so they are not merely different in degree but only partially aligned.

Besides, each stage's model weights visibly move toward the corresponding oracle, confirming that every stage does pull the weights in its intended task direction. However, because these weights are simultaneously shaped by multiple stages of training, no stage aligns perfectly with its oracle. 
The more stages a checkpoint has passed through, the further its starting point has already drifted from Base, and the harder it becomes to recover the single task solution, direct visual evidence that accumulated cross-stage interference, rather than the difficulty of any single task, is what prevents alignment. More supportive evidence on checkpints' cosine similarity matrix of different stages is shown in Appendix \ref{app:model-level}.

\subsubsection{\textbf{Token-level Diagnosis: Prediction Stability across Stages}}
To localize where inside a generation the model is rewritten, Figure~\ref{fig:d} fixes one set of token ids decoded by \textsc{Seq-Math} and forces every subsequent checkpoint to score that same sequence, so that all checkpoints are compared at identical positions. At each position, we record whether the checkpoint still ranks \textsc{Seq-Math}'s token first; the token-level prediction consistency of a group of positions is the fraction at which it does. Splitting positions by \textsc{Seq-Math}'s predictive entropy separates two very different behaviors. The overwhelming majority of positions are low-entropy, the model was already committed to a single continuation, and they survive the entire sequence of training stages essentially untouched, while almost all of the change is carried by the small high-entropy minority. 
Sequential training does not alter token predictions uniformly. Under fixed-prefix evaluation, low-entropy positions remain highly stable across subsequent stages, whereas prediction changes are disproportionately concentrated at positions where the reference checkpoint assigns higher uncertainty.
Details of the analysis are in Appendix~\ref{app:token-level}.
\begin{finding}[label=fin:interference]{Tasks induce partially aligned parameter updates}
Task-specific training moves the model toward distinct directions, while sequential training reconciles these partially misaligned updates, leading to uneven transfer and forgetting across tasks.
\end{finding}

\section{Method}
We present three complementary technical routes for model integration from different dimensions, as shown in Figure \ref{fig:main}.

\subsection{Multi-teacher On-policy Distillation}

\noindent\textbf{Vanilla OPD~\citep{lu2025onpolicydistillation}}.
For a single stage transition from $\pi_i$ to $\pi_{i+1}$, we regard the previous-stage policy $\pi_i$ as the teacher policy and optimize the student policy $\pi_{i+1}$ via OPD. 
Specifically, trajectories are sampled from $\pi_{i+1}$ under the stage-specific prompt distribution $\mathcal{D}_i$, while the teacher $\pi_i$ provides token-level supervision on these student-generated trajectories.
The objective based on reverse KL minimizes the token-level divergence on student-generated rollouts,
\begin{equation}
\mathcal{L}_{\mathrm{reverse}}^{(i)}
=
\mathbb{E}_{x\sim\mathcal{D}_i,\,
\hat y\sim\pi_{i+1}}
\left[
\sum_{t=1}^{T}
D_{\mathrm{KL}}
\!\left(
\pi_{i+1}(\cdot\mid h_t)
\;\|\;
\pi_{i}(\cdot\mid h_t)
\right)
\right],
\end{equation}

where $h_t=(x,\hat y_{<t})$ denotes the context at step $t$. We write $\mathcal{L}_{\mathrm{reverse},t}^{(i)}=D_{\mathrm{KL}}\!\left(\pi_{i+1}(\cdot\mid h_t)\;\|\;\pi_{i}(\cdot\mid h_t)\right)$ for the per-position reverse-KL term inside the sum, so that $\mathcal{L}_{\mathrm{reverse}}^{(i)}=\mathbb{E}_{x\sim\mathcal{D}_i,\,\hat y\sim\pi_{i+1}}\big[\sum_{t=1}^{T}\mathcal{L}_{\mathrm{reverse},t}^{(i)}\big]$. The strengths and limitations of reverse KL arise from the same underlying property. Because its expectation is taken exclusively over student-generated trajectories, reverse KL efficiently penalizes student outputs that receive low probability while providing no learning signal for teacher modes that the student rarely explores. This mode-seeking behavior can progressively concentrate the student distribution on a subset of high-probability modes, causing the student to even lose the teacher’s diverse reasoning paths. Forward KL~\citep{agarwal2024policy} mitigates this issue through its mode-covering behavior, encouraging the student to retain a broader support of the teacher distribution. However, this benefit comes at a cost: forward KL requires the student to account for low-probability tail tokens, increasing computational and memory overhead. More importantly, when the student has limited capacity, aggressively covering the teacher’s full distribution may spread its probability mass across less informative modes.

\noindent\textbf{Mixed OPD (MOPD).}
We argue that the training collapse observed on agentic search tasks (Appendix~\ref{sec:supp_mopd}) is a structural consequence of the reverse-KL objective. On low-entropy tokens, this asymmetry is exactly what we want; the teacher is near-deterministic, and sharp imitation is appropriate. On high-entropy tokens, however, which in agentic trajectories correspond precisely to decision points, \textit{e.g.,} choosing among alternative search queries or tool invocations, the teacher's uncertainty encodes a set of comparably viable strategies. Under mode-seeking pressure, the student commits to a single branch and progressively prunes the rest, manifesting as entropy collapse and eventual divergence. The failure is thus token-selective, and so should be the remedy: we retain reverse KL wherever imitation ought to be sharp, and inject a mass-covering forward-KL term only where the teacher itself is uncertain:
\begin{equation}
\label{eq:eopd}
\mathcal{L}_{\mathrm{MOPD},t}^{(i)}
=
\mathcal{L}_{\mathrm{reverse},t}^{(i)}
+
\gamma\,
\mathbf{1}\!\left(H_t^{\mathrm{tea}}>\alpha\right)
\sum_{v \in \mathcal{S}_t^{q}}
\tilde{\pi}_{i}(v \mid h_t)\,
\log
\frac{\tilde{\pi}_{i}(v \mid h_t)}
{\pi_{i+1}(v \mid h_t)},
\end{equation}
where $H_t^{\mathrm{tea}}$ is the token-level entropy of the teacher $\pi_i$, $\alpha$ is an entropy threshold, and $\tilde{\pi}_{i}(\cdot\mid h_t)$ is the teacher distribution renormalized over its top-$q$ token set $\mathcal{S}_t^{q}$, \textit{i.e.,} $\tilde{\pi}_{i}(v\mid h_t)\propto\pi_{i}(v\mid h_t)$ for $v\in\mathcal{S}_t^{q}$ and zero otherwise, where $v$ ranges over vocabulary tokens. Since the teacher's logits are available at distillation time, the forward-KL term is evaluated in closed form over $\mathcal{S}_t^{q}$ rather than by Monte-Carlo sampling; the top-$q$ truncation simultaneously shields the student from noisy supervision in the teacher's low-probability tail and keeps the overhead negligible~\citep{shum2024first, peng2025pre}.

\noindent\textbf{Multi-teacher Mixed OPD (MMOPD).}
Since we have the full $K$-stage curriculum, we further extend it to a multi-teacher setting. Let $\{\pi_k\}_{k=1}^{K}$ denote a set of frozen teacher policies. The student $\pi_\theta$ is initialized from the last checkpoint $\pi_K$ and trained on a mixed prompt stream drawn from all stage-specific distributions with balanced proportions, $\mathcal{D}_{\mathrm{mix}} = \tfrac{1}{K}\sum_{k}\mathcal{D}_k$. Balancing is necessary because the raw prompt pools differ by orders of magnitude in size; an unbalanced mixture would let a single domain dominate every batch and starve the remaining teachers of gradient signal.

Each prompt carries a domain tag $d(x)\in\{1,\ldots,K\}$ assigned earlier, which will be used for the routing of teachers. At rollout time, the student interacts under the native protocol of its domain, the resulting trajectory is scored exclusively by the domain-owning teacher $\pi_{d(x)}$, which performs a single scoring-only forward pass over the student-generated tokens to expose the per-token signal required by Eq.~\eqref{eq:eopd}. We deliberately route rather than ensemble: averaging teacher logits across domains is ill-posed, as a teacher specialized in one interaction protocol provides no meaningful supervision on trajectories from another. Tokens emitted by the environment, such as tool observations and simulated-user turns, are masked out of the objective. The objective is as follows:
\begin{equation}
\label{eq:mopd}
\mathcal{L}_{\mathrm{MMOPD}}
=
\mathbb{E}_{k\sim\mathcal{U}(K)}\;
\mathbb{E}_{x\sim\mathcal{D}_k,\,\hat{y}\sim\pi_{\theta}}
\left[
\sum_{t=1}^{T}
\mathcal{L}_{\mathrm{MOPD},t}\!\left(\pi_{\theta}\,;\,\pi_{k}\right)
\right],
\end{equation}
Since all teachers are frozen and invoked only for scoring, the additional cost over single-teacher OPD is a constant number of forward passes per trajectory, independent of the number of capabilities being consolidated.

\subsection{Self-distilled Fine-tuning}

The core idea of SDFT is to use high-quality reasoning trajectories generated by the oracle model itself as supervision signals, iteratively reinforcing its existing capabilities through self-distillation.

\noindent\textbf{Oracle construction.} For each stage-specific task $k \in \{1,\ldots,K\}$, we depart from the base policy $\pi_0$ and train a dedicated oracle policy $\pi_k^{\star}$ using the same task-specific prompt distribution $\mathcal{D}_k$ and reward $r_k$ as above, but without any cross-stage dependency:
\[
\pi_0 \overset{(\mathcal{D}_k,\,r_k)}{\longrightarrow} \pi_k^{\star}, \qquad k \in \{ 1,\ldots,K \}.
\]
Because each $\pi_k^{\star}$ is optimized in isolation, it represents the near-upper-bound capability attainable on task $k$ under our pipeline, free from the gradient interference that a shared trajectory incurs. This oracle set $\{\pi_k^{\star}\}_{k=1}^{K}$ serves as the source of the highest-fidelity behavioral demonstrations for each capability.

\noindent\textbf{Trajectory filtering.} We then roll out each oracle $\pi_k^{\star}$ on its corresponding prompt distribution $\mathcal{D}_k$ and retain only trajectories that meet stringent quality criteria. Concretely, we filter on final-reward correctness, well-formed tool invocation, and interaction-protocol compliance, and additionally deduplicate near-identical trajectories to preserve behavioral diversity. Let $\mathcal{T}_k = \{(x,\hat{y}) : x \sim \mathcal{D}_k,\; \hat{y} \sim \pi_k^{\star}(\cdot \mid x),\; \phi(\hat{y}) = 1\}$ denote the filtered high-quality set of prompt--trajectory pairs, where $\phi(\cdot)$ is the composite acceptance indicator. The consolidated student $\pi_{K}$ is subsequently obtained by sequential training:
\begin{equation}
\label{eq:sdft}
\mathcal{L}_{\mathrm{SDFT}}
=
-\,\mathbb{E}_{k\sim\mathcal{P}}\;
\mathbb{E}_{(x,\hat y)\sim\mathcal{T}_k}
\left[
\sum_{t=1}^{T}
\log \pi_{K}\!\left(\hat y_t \mid h_t\right)
\right],
\end{equation}
where $\mathcal{P}$ denotes the sampling distribution over tasks that governs the data mixing ratio across capabilities.

\noindent\textbf{Data composition and ordering.} The effectiveness of SDFT is highly sensitive to two factors that are absent in the RL stages: the mixing ratio across domains and the presentation order of the training data. An imbalanced mixture allows high-volume tasks to dominate the gradient and suppress under-represented capabilities, whereas a strictly ordered schedule reintroduces the very forgetting dynamics that motivated this paradigm. We therefore conduct extensive ablations over the task-mixing distribution $\mathcal{P}$ and over curriculum orderings, ranging from fully interleaved shuffling to staged and difficulty-graded schedules, to identify the configuration that best balances all capabilities in the consolidated model.

\begin{table*}[t]
\centering
\caption{\textbf{Main result.}
Each per-task expert is trained independently on a single task.
\textsc{Seq-Final} denotes the final checkpoint obtained after sequentially training.
Starting from this checkpoint, we further apply \textsc{MMOPD} and \textsc{SDFT}.
\textsc{MM} uniformly averages the parameters.
Dashed boxes indicate the training stage corresponding to each task.}
\vspace{-0.2cm}
\label{tab:main_mopd}
\setlength{\tabcolsep}{3.5pt}
\renewcommand{\arraystretch}{1.05}

\resizebox{\textwidth}{!}{%
\begin{tabular}{l cccc ccccccc}
\toprule
& \multicolumn{4}{c}{\textbf{In-domain evaluation} ($\uparrow$)}
& \multicolumn{7}{c}{\textbf{Out-of-domain evaluation} ($\uparrow$)} \\
\cmidrule(lr){2-5}
\cmidrule(lr){6-12}

& \multirow{2}{*}{\shortstack{AIME26\\(avg@16)}}
& \multirow{2}{*}{NQ}
& \multirow{2}{*}{$\tau^3$-Retail}
& \multirow{2}{*}{IF-Eval}
& \multicolumn{2}{c}{Math}
& \multicolumn{1}{c}{Single-hop Search}
& \multicolumn{1}{c}{Multi-hop Search}
& \multicolumn{2}{c}{E-commerce}
& \multicolumn{1}{c}{IF} \\

\cmidrule(lr){6-7}
\cmidrule(lr){8-8}
\cmidrule(lr){9-9}
\cmidrule(lr){10-11}
\cmidrule(lr){12-12}

\textbf{Pipeline}
& & & &
& GPQA & MMLU
& Avg.
& Avg.
& $\tau^3$-Telecom & $\tau^3$-Mock
& IF-Bench \\

\midrule

\rowcolor{uclablue}
\multicolumn{12}{l}{
    \textcolor{white}{\emph{Per-Task Oracle Training}}
} \\

\midrule

Expert$_{\text{Math}}$
& \dashbox{$\textcolor{red}{\mathbf{25.83}}$}
& $22.0_{\scriptscriptstyle \pm 0.2}$
& $2.3_{\scriptscriptstyle \pm 0.5}$
& $43.3_{\scriptscriptstyle \pm 0.6}$
& \dashbox{$40.2_{\scriptscriptstyle \pm 2.5}$}
& \dashbox{$79.0_{\scriptscriptstyle \pm 0.2}$}
& $38.7_{\scriptscriptstyle \pm 0.2}$
& $22.8_{\scriptscriptstyle \pm 0.5}$
& $13.1_{\scriptscriptstyle \pm 2.1}$
& $3.3_{\scriptscriptstyle \pm 3.8}$
& $17.1_{\scriptscriptstyle \pm 0.2}$ \\

Expert$_{\text{Search}}$
& $9.79$
& \dashbox{$\textcolor{red}{\mathbf{49.9_{\scriptscriptstyle \pm 1.7}}}$}
& $2.7_{\scriptscriptstyle \pm 0.2}$
& $62.4_{\scriptscriptstyle \pm 0.2}$
& $40.2_{\scriptscriptstyle \pm 1.9}$
& $79.2_{\scriptscriptstyle \pm 0.1}$
& \dashbox{$\underline{{56.0_{\scriptscriptstyle \pm 1.3}}}$}
& \dashbox{${36.7_{\scriptscriptstyle \pm 2.3}}$}
& $6.4_{\scriptscriptstyle \pm 0.7}$
& $36.7_{\scriptscriptstyle \pm 10.1}$
& $14.7_{\scriptscriptstyle \pm 0.3}$ \\

Expert$_{\text{E-commerce}}$
& $6.25$
& $19.3_{\scriptscriptstyle \pm 0.1}$
& \dashbox{$\textcolor{red}{\mathbf{33.2_{\scriptscriptstyle \pm 1.1}}}$}
& $40.1_{\scriptscriptstyle \pm 0.4}$
& $33.2_{\scriptscriptstyle \pm 2.6}$
& $70.1_{\scriptscriptstyle \pm 0.1}$
& $32.2_{\scriptscriptstyle \pm 0.2}$
& $19.6_{\scriptscriptstyle \pm 0.2}$
& \dashbox{$\textcolor{red}{\mathbf{46.8_{\scriptscriptstyle \pm 1.3}}}$}
& \dashbox{$65.8_{\scriptscriptstyle \pm 1.3}$}
& $15.7_{\scriptscriptstyle \pm 0.3}$ \\

Expert$_{\text{IF}}$
& $0.00$
& $26.4_{\scriptscriptstyle \pm 1.1}$
& $4.6_{\scriptscriptstyle \pm 0.2}$
& \dashbox{$\textcolor{red}{\mathbf{86.2_{\scriptscriptstyle \pm 0.0}}}$}
& $39.0_{\scriptscriptstyle \pm 1.8}$
& $\underline{80.2_{\scriptscriptstyle \pm 0.2}}$
& $45.1_{\scriptscriptstyle \pm 1.6}$
& $25.3_{\scriptscriptstyle \pm 2.5}$
& $8.9_{\scriptscriptstyle \pm 2.0}$
& $9.2_{\scriptscriptstyle \pm 1.4}$
& \dashbox{$29.0_{\scriptscriptstyle \pm 0.0}$} \\

\midrule

\rowcolor{uclablue}
\multicolumn{12}{l}{
    \textcolor{white}{\emph{Multi-Stage Post-Training}}
} \\

Seq-${\text{Final}}$
& $10.21$
& $33.5_{\scriptscriptstyle \pm 1.7}$
& ${29.6_{\scriptscriptstyle \pm 1.6}}$
& ${84.8_{\scriptscriptstyle \pm 0.1}}$
& $37.9_{\scriptscriptstyle \pm 1.7}$
& $79.5_{\scriptscriptstyle \pm 0.0}$
& $44.9_{\scriptscriptstyle \pm 0.7}$
& $25.0_{\scriptscriptstyle \pm 1.5}$
& $\underline{45.9_{\scriptscriptstyle \pm 1.8}}$
& $62.5_{\scriptscriptstyle \pm 6.6}$
& $\textcolor{red}{\mathbf{30.8_{\scriptscriptstyle \pm 0.2}}}$ \\

\quad +MMOPD
& $21.25$
& $45.2_{\scriptscriptstyle \pm 0.4}$
& $27.7_{\scriptscriptstyle \pm 1.5}$
& $84.6_{\scriptscriptstyle \pm 0.5}$
& $40.1_{\scriptscriptstyle \pm 1.2}$
& $79.2_{\scriptscriptstyle \pm 0.2}$
& $53.7_{\scriptscriptstyle \pm 0.0}$
& $32.8_{\scriptscriptstyle \pm 0.1}$
& $\underline{45.9_{\scriptscriptstyle \pm 2.5}}$
& $\underline{66.3_{\scriptscriptstyle \pm 1.8}}$
& ${30.5_{\scriptscriptstyle \pm 0.2}}$ \\

\quad +SDFT
& $\underline{22.29}$
& ${48.3_{\scriptscriptstyle \pm 0.1}}$
& $22.3_{\scriptscriptstyle \pm 0.5}$
& $53.6_{\scriptscriptstyle \pm 0.2}$
& $39.7_{\scriptscriptstyle \pm 3.7}$
& $78.9_{\scriptscriptstyle \pm 0.1}$
& ${55.6_{\scriptscriptstyle \pm 0.1}}$
& $\underline{{38.0_{\scriptscriptstyle \pm 0.2}}}$
& $21.7_{\scriptscriptstyle \pm 1.7}$
& $65.0_{\scriptscriptstyle \pm 2.5}$
& $19.7_{\scriptscriptstyle \pm 0.0}$ \\

\quad +MM
& $14.79$
& $39.0_{\scriptscriptstyle \pm 0.2}$
& $11.8_{\scriptscriptstyle \pm 1.9}$
& $74.6_{\scriptscriptstyle \pm 0.1}$
& $\underline{41.8_{\scriptscriptstyle \pm 0.9}}$
& $\textcolor{red}{\mathbf{80.3_{\scriptscriptstyle \pm 0.2}}}$
& $51.0_{\scriptscriptstyle \pm 0.0}$
& $36.2_{\scriptscriptstyle \pm 0.4}$
& $16.5_{\scriptscriptstyle \pm 0.3}$
& $27.5_{\scriptscriptstyle \pm 0.0}$
& $20.0_{\scriptscriptstyle \pm 0.0}$ \\

\midrule

MLE (Ours)
& $21.04$
& $\underline{49.7_{\scriptscriptstyle \pm 0.2}}$
& $\underline{32.9_{\scriptscriptstyle \pm 2.2}}$
& $\underline{85.0_{\scriptscriptstyle \pm 0.3}}$
& $\textcolor{red}{\mathbf{42.4_{\scriptscriptstyle \pm 1.6}}}$
& $79.9_{\scriptscriptstyle \pm 0.2}$
& $\textcolor{red}{\mathbf{57.7_{\scriptscriptstyle \pm 0.0}}}$
& $\textcolor{red}{\mathbf{38.6_{\scriptscriptstyle \pm 0.1}}}$
& $45.6_{\scriptscriptstyle \pm 2.4}$
& $\textcolor{red}{\mathbf{66.7_{\scriptscriptstyle \pm 1.4}}}$
& $\underline{29.1_{\scriptscriptstyle \pm 0.2}}$ \\

\bottomrule
\end{tabular}
\vspace{-0.5cm}
}
\end{table*}
\subsection{Model Merging}

Given a set of specialized models that each master a single capability, it directly combines their weights without any further gradient updates or trajectory generation. This paradigm is appealing because it is training-free at the consolidation step.
We reuse the sequential checkpoint set $\{\pi_k\}_{k=1}^{K}$ constructed above. Since all of them are fine-tuned from a common initialization, their parameters remain linearly connected in a shared region of weight space, which makes direct weighted averaging meaningful. The merged model is obtained as:
\begin{equation}
\theta_{\mathrm{merge}}
=
\theta_0
+
\sum_{k=1}^{K}
\lambda_k\,\bigl(\theta_k-\theta_0\bigr),
\qquad
\lambda_k \ge 0,
\end{equation}
where $\tau_k = \theta_k-\theta_0$ is the task vector that captures the parameter displacement induced by specializing in task $k$, and $\lambda_k$ is its associated interpolation weight. Setting $\sum_k \lambda_k = 1$ recovers a convex combination of these model weights, while relaxing this constraint permits amplifying or attenuating individual task vectors to counteract capability imbalance in the merged model.

\noindent\textbf{Weighting schemes.} The behavior of the merged model is governed entirely by the coefficient vector $\boldsymbol{\lambda} = (\lambda_1,\ldots,\lambda_K)$. We explore a spectrum of weighting schemes to characterize the trade-off surface among capabilities: (i) uniform averaging, where $\lambda_k = 1/K$ for all $k$, treating every capability as equally important; (ii) grid-searched weighting, where the coefficients are tuned to maximize aggregate validation performance across all tasks; and (iii) capability-balanced weighting, where under-performing capabilities receive larger coefficients to compensate for their weaker task vectors. Because each capability responds differently to the magnitude of its task vector, the merged performance is highly non-uniform in $\boldsymbol{\lambda}$.






\section{Exploration of Different ACL Paradigms}

\subsection{Experiment Setup}
\noindent\textbf{Implementation Details.}
We construct one per-task oracle expert by applying each task-specific training pipeline independently to Qwen3-8B-Base~\citep{qwen3technicalreport}. For \textsc{SDFT}, each expert performs one rollout pass over its training set with group size 8, and the resulting trajectories are combined to fine-tune the final sequential checkpoint. For \textsc{MMOPD}, the four sequential-stage checkpoints serve as task-specific teachers and the final sequential checkpoint as the student. For \textsc{MM}, we uniformly average the four sequential-stage checkpoints. Our implementation is built on the \textsc{Slime}~\citep{slime_github} RL framework with an agentic loop for rollouts and environment interaction. All experiments use two NVIDIA H200 nodes; additional hyperparameters are provided in the Appendix.

\noindent\textbf{Evaluation Metrics.}
We evaluate all models on in-domain tasks seen during post-training and out-of-domain tasks that measure cross-domain generalization. Math performance is reported on AIME 2026~\citep{dekoninck2026matharena}, GPQA-Diamond~\citep{rein2023gpqagraduatelevelgoogleproofqa}, and MMLU-Redux~\citep{gema2025mmlu}. For search, NQ~\citep{kwiatkowski-etal-2019-natural} is the in-domain benchmark; out-of-domain results are averaged over PopQA~\citep{mallen2023llm_memorization} and TriviaQA~\citep{joshi2017triviaqalargescaledistantly} for single-hop search, and 2WikiMultiHopQA~\citep{xanh2020_2wikimultihop}, Bamboogle~\citep{press2023measuringnarrowingcompositionalitygap}, HotpotQA~\citep{yang2018hotpotqadatasetdiverseexplainable}, and MuSiQue~\citep{trivedi2022musiquemultihopquestionssinglehop} for multi-hop search. E-commerce is evaluated by task success rate on the $\tau^3$-Bench retail domain for in-domain evaluation and the telecom and mock domains for out-of-domain evaluation~\citep{yao2024taubenchbenchmarktoolagentuserinteraction,barres2025tau2}. Instruction following is evaluated on IF-Eval~\citep{zhou2023instructionfollowingevaluationlargelanguage} for in-domain evaluation and IF-Bench~\citep{pyatkin2025generalizing} for out-of-domain evaluation. Further details are provided in Appendix~\ref{appendix_eval_detial}.

\subsection{Main Results}

Table~\ref{tab:main_mopd} highlights the difficulty of consolidating heterogeneous post-training capabilities within a single shared model. Detailed setting can be seen in Table~\ref{tab:model_overview}. The independently trained per-task oracle exhibits clear specialization, but their improvements do not transfer consistently across domains. For example, the search expert achieves 49.9 on NQ and 56.0 on single-hop search, while the IF expert reaches 86.2 on IF-Eval. However, both models perform substantially worse on several unrelated tasks. Sequential training accumulates capabilities across stages, but later updates modify or overwrite behaviors acquired earlier. The final checkpoint retains strong performance on later-stage tasks, including 45.9 on $\tau^3$-Telecom and 30.8 on IF-Bench, while its AIME26 score falls to 10.21 and its NQ score remains well below the search expert’s 49.9. This pattern indicates substantial cross-task interference and forgetting rather than uniform capability accumulation.

Post-hoc consolidation methods recover different subsets of these capabilities, but none improves the shared checkpoint consistently across all domains. MMOPD provides a relatively favorable trade-off: compared with Sequential, it improves AIME26 from 10.21 to 21.25, NQ from 33.5 to 45.2, single-hop search from 44.9 to 53.7, and multi-hop search from 25.0 to 32.8. It also largely preserves IF performance. Nevertheless, these gains are accompanied by slightly lower performance on MMLU and IF-Bench. SDFT shows a different trade-off, recovering search capabilities more strongly, reaching 48.3 on NQ, 55.6 on single-hop search, and 38.0 on multi-hop search, but substantially degrading E-commerce and IF performance. Model merging achieves the strongest MMLU result of 80.3, yet performs worse than Sequential or MMOPD on most E-commerce and IF benchmarks. Overall, these results suggest that the problem cannot be solved simply by finding a better shared checkpoint: methods that recover one capability often interfere with others, motivating mechanisms that preserve task-specific parameters rather than forcing all capabilities into a single parameter space.

\begin{finding}[label=fin:consolidation]{Shared-model consolidation causes trade-offs}
Methods that recover one capability often degrade others, indicating persistent cross-task interference in a single shared model.
\end{finding}

\begin{table*}[t]
\centering
\caption{
Ablation on single-teacher distillation. The teacher is the search-stage checkpoint from sequential training, and the student is the subsequent E-commerce-stage checkpoint. Student$_{\text{MOPD}}$ is obtained by distilling the search-stage teacher into the student. Dashed boxes indicate the training stage corresponding to each task. 
}
\vspace{-0.3cm}
\label{tab:distill}
\setlength{\tabcolsep}{3.5pt}
\renewcommand{\arraystretch}{1.05}
\resizebox{\textwidth}{!}{%
\begin{tabular}{l cccc ccccccc}
\toprule
& \multicolumn{4}{c}{\textbf{In-domain evaluation} ($\uparrow$)}
& \multicolumn{7}{c}{\textbf{Out-of-domain evaluation} ($\uparrow$)} \\
\cmidrule(lr){2-5}
\cmidrule(lr){6-12}
& \multirow{2}{*}{\shortstack{AIME26\\(avg@16)}}
& \multirow{2}{*}{NQ}
& \multirow{2}{*}{$\tau^3$-Retail}
& \multirow{2}{*}{IF-Eval}
& \multicolumn{2}{c}{Math}
& \multicolumn{1}{c}{Single-hop Search}
& \multicolumn{1}{c}{Multi-hop Search}
& \multicolumn{2}{c}{E-commerce}
& \multicolumn{1}{c}{IF} \\

\cmidrule(lr){6-7}
\cmidrule(lr){8-8}
\cmidrule(lr){9-9}
\cmidrule(lr){10-11}
\cmidrule(lr){12-12}

\textbf{Pipeline}
& & & &
& GPQA & MMLU
& Avg.
& Avg.
& $\tau^3$-Telecom & $\tau^3$-Mock
& IF-Bench \\

\midrule

\rowcolor{uclablue}
\multicolumn{12}{l}{
    \textcolor{white}{\emph{Single-teacher distillation}
    (Student: Seq-${\text{E-commerce}}$ $\leftarrow$ Teacher: Seq-${\text{Search}}$)}
} \\

\midrule

Teacher$_{\text{Search}}$
& $\textcolor{red}{\mathbf{23.33}}$
& \dashbox{$45.2_{\scriptscriptstyle \pm 0.4}$}
& $2.0_{\scriptscriptstyle \pm 0.3}$
& $39.8_{\scriptscriptstyle \pm 0.2}$
& $\textcolor{red}{\mathbf{37.2_{\scriptscriptstyle \pm 0.7}}}$
& $\textcolor{red}{\mathbf{79.2_{\scriptscriptstyle \pm 0.1}}}$
& \dashbox{$52.9_{\scriptscriptstyle \pm 0.1}$} 
& \dashbox{$37.4_{\scriptscriptstyle \pm 0.8}$}
& $9.1_{\scriptscriptstyle \pm 3.9}$
& $30.8_{\scriptscriptstyle \pm 1.4}$
& $15.8_{\scriptscriptstyle \pm 0.2}$ \\

Student$_{\text{E-commerce}}$
& $6.04$
& $14.6_{\scriptscriptstyle \pm 0.1}$
& \dashbox{$\textcolor{red}{\mathbf{34.0_{\scriptscriptstyle \pm 0.9}}}$}
& $41.3_{\scriptscriptstyle \pm 0.3}$
& $36.0_{\scriptscriptstyle \pm 1.4}$
& $67.3_{\scriptscriptstyle \pm 0.3}$
& $13.2_{\scriptscriptstyle \pm 0.1}$
& $9.4_{\scriptscriptstyle \pm 0.2}$
& \dashbox{$\textcolor{red}{\mathbf{46.3_{\scriptscriptstyle \pm 1.6}}}$}
& \dashbox{$\textcolor{red}{\mathbf{74.2_{\scriptscriptstyle \pm 1.4}}}$}
& $19.3_{\scriptscriptstyle \pm 0.9}$ \\

\midrule

Student$_{\text{MOPD}}$
& $15.62$
& $\textcolor{red}{\mathbf{48.4_{\scriptscriptstyle \pm 0.0}}}$
& $15.8_{\scriptscriptstyle \pm 0.6}$
& $\textcolor{red}{\mathbf{53.6_{\scriptscriptstyle \pm 0.2}}}$
& $35.1_{\scriptscriptstyle \pm 1.4}$
& $78.7_{\scriptscriptstyle \pm 0.1}$
& $\textcolor{red}{\mathbf{53.8_{\scriptscriptstyle \pm 0.2}}}$
& $\textcolor{red}{\mathbf{39.0_{\scriptscriptstyle \pm 0.2}}}$
& $21.7_{\scriptscriptstyle \pm 1.4}$
& $65.0_{\scriptscriptstyle \pm 2.0}$
& $\textcolor{red}{\mathbf{19.7_{\scriptscriptstyle \pm 0.0}}}$ \\

\bottomrule
\end{tabular}%
}
\end{table*}

\begin{table*}[t]
\centering
\caption{
Comparison of SDFT variants with different initializations and data mixing strategies. SDFT is initialized either from the sequentially trained checkpoint ($\pi_K$) or the base model ($\pi_0$), using oracle trajectories collected from the corresponding policy. Random-Mix samples training examples randomly, whereas Balanced-Mix enforces uniform sampling.
}
\vspace{-0.3cm}
\label{tab:sftvariants}
\setlength{\tabcolsep}{3.5pt}
\renewcommand{\arraystretch}{1.05}
\resizebox{\textwidth}{!}{%
\begin{tabular}{l cccc ccccccc}
\toprule
& \multicolumn{4}{c}{\textbf{In-domain evaluation} ($\uparrow$)}
& \multicolumn{7}{c}{\textbf{Out-of-domain evaluation} ($\uparrow$)} \\
\cmidrule(lr){2-5}
\cmidrule(lr){6-12}

& \multirow{2}{*}{\shortstack{AIME26\\(avg@16)}}
& \multirow{2}{*}{NQ}
& \multirow{2}{*}{$\tau^3$-Retail}
& \multirow{2}{*}{IF-Eval}
& \multicolumn{2}{c}{Math}
& \multicolumn{1}{c}{Single-hop Search}
& \multicolumn{1}{c}{Multi-hop Search}
& \multicolumn{2}{c}{E-commerce}
& \multicolumn{1}{c}{IF} \\

\cmidrule(lr){6-7}
\cmidrule(lr){8-8}
\cmidrule(lr){9-9}
\cmidrule(lr){10-11}
\cmidrule(lr){12-12}

\textbf{Pipeline}
& & & &
& GPQA & MMLU
& Avg.
& Avg.
& $\tau^3$-Telecom & $\tau^3$-Mock
& IF-Bench \\

\midrule

\rowcolor{uclablue}
\multicolumn{12}{l}{
    \textcolor{white}{Seq + SDFT}
} \\

\midrule

Random-Mix
& $\textcolor{red}{\mathbf{22.92}}$
& $48.3$
& $\textcolor{red}{\mathbf{25.0}}$
& $54.0$
& $\textcolor{red}{\mathbf{44.4}}$
& $79.2$
& $55.7$
& $37.0$
& $\textcolor{red}{\mathbf{23.9}}$
& $65.0$
& $\textcolor{red}{\mathbf{21.3}}$ \\

Balanced-Mix
& $22.29$
& $\textcolor{red}{\mathbf{48.4}}$
& $22.3$
& $53.8$
& $38.26$
& $78.86$
& $55.6$
& $\textcolor{red}{\mathbf{38.2}}$
& $20.0$
& $\textcolor{red}{\mathbf{67.5}}$
& $19.7$ \\

\midrule

\rowcolor{uclablue}
\multicolumn{12}{l}{
    \textcolor{white}{Base + SDFT}
} \\

\midrule

Random-Mix
& $20.83$
& $48.2$
& $5.9$
& $\textcolor{red}{\mathbf{61.6}}$
& $39.6$
& $\textcolor{red}{\mathbf{79.7}}$
& $\textcolor{red}{\mathbf{55.9}}$
& $36.4$
& $2.6$
& $42.5$
& $19.7$ \\

Balanced-Mix
& $20.00$
& $48.0$
& $10.1$
& $57.3$
& $43.6$
& $79.5$
& $55.4$
& $\textcolor{red}{\mathbf{38.2}}$
& $6.6$
& $\textcolor{red}{\mathbf{67.5}}$
& $21.0$ \\

\bottomrule
\end{tabular}%
}
\end{table*}

\subsection{Ablation Study}
\subsubsection{\textbf{Single-teacher OPD}}
To verify that our proposed Mixed OPD is effective on both reasoning and agentic tasks, we first select the Search task, which exhibits the highest degree of forgetting, for single-teacher OPD. Table~\ref{tab:distill} clearly demonstrates that the student's Search capability improves substantially after OPD, even surpassing the original Search teacher. This indicates that success on the agentic Search task lies not only in correctly invoking tools, but also in performing more advanced reasoning to make judgments. Returning to OPD after training on other tasks is therefore a highly meaningful choice: it not only substantially restores the forgotten search capability, but also largely preserves the E-commerce capability. More ablations are shown in Appendix~\ref{sec:supp_mopd}.

\subsubsection{\textbf{SDFT Replay Variant}}
We compare the effects of SDFT across different models, in order to infer whether SDFT is suitable as a solution for consolidating multiple capabilities. In Table~\ref{tab:sftvariants}, we find that starting from $\pi_K$, the model obtained from sequential training, yields better overall learning, because it has already traversed the full sequence and thus carries its own capability bias into the SFT data. Specifically, since math and IF are foundational capabilities, the differences there are marginal, whereas the gap becomes far more pronounced on the two agentic tasks. We therefore adopt it as our primary SDFT method.

\subsubsection{\textbf{SDFT Data Recipe}}
We find that the data mixing strategy, \textit{i.e.,} the task sampling distribution $\mathcal{P}$ in Eq.~\eqref{eq:sdft}, also remains important. Specifically, we compare Random-Mix and Balanced-Mix: the former samples training examples at random from the pooled data, so that each task is represented in proportion to the size of its trajectory set, whereas the latter enforces strictly uniform sampling across tasks. During training, we observe that balanced sampling stabilizes the optimization gradients arising from the differing directions of the multiple tasks, leading to faster and lower loss reduction. The results in Table~\ref{tab:sftvariants} likewise show that Balanced-Mix achieves better performance on average.

\subsection{Analysis \& Findings}
\begin{figure}[t]
\centering
\begin{minipage}{0.85\linewidth}
\centering
\begin{subfigure}[t]{0.52\linewidth}
    \centering
    \includegraphics[width=\linewidth]{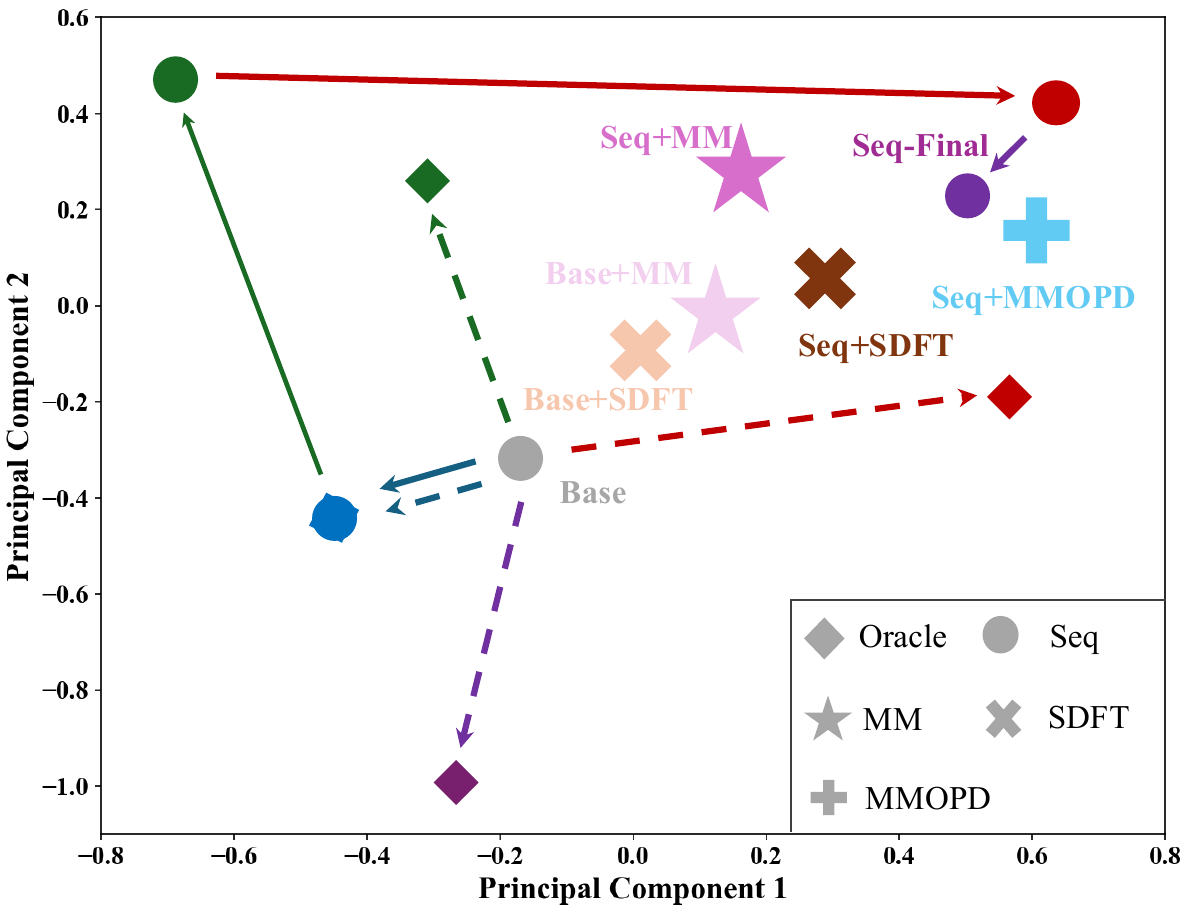}
    \caption{Principal component analysis of model weights in different methods}
    \label{fig:pca_analysis}
\end{subfigure}%
\hfill
\begin{subfigure}[t]{0.42\linewidth}
    \centering
    \includegraphics[width=\linewidth]{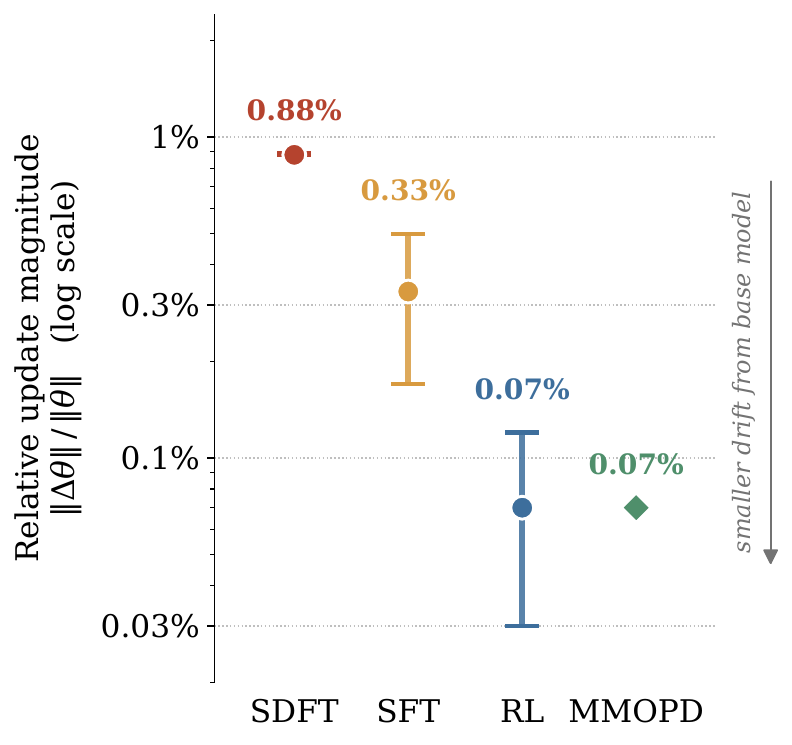}
    \caption{Relative weight update magnitude of different pipelines}
    \label{fig:magnitude}
\end{subfigure}
\end{minipage}
\vspace{-0.3cm}
\caption{Magnitude and direction of parameter updates across post-training pipelines. (a) PCA of model weights. MMOPD remains close to the final sequential checkpoint, whereas SDFT moves farther in parameter space. (b) Relative update magnitude ($\|\Delta\theta\|/\|\theta\|$). SFT-based methods produce much larger parameter updates than RL-based methods, with SDFT exhibiting the largest changes.}
\vspace{-0.5cm}
\end{figure}
\subsubsection{Magnitude and Direction of Different Post-Training Pipeline}
We next analyze how different post-training pipelines modify the model in both magnitude and direction. Figure~\ref{fig:magnitude} reports the relative parameter-update magnitude, $\|\Delta\theta\|/\|\theta\|$. For \textsc{SDFT} and \textsc{MMOPD}, we measure the update relative to the final sequential checkpoint. For task-specific SFT, we analyze search and E-commerce updates under both oracle and sequential training. For RL, we aggregate updates across all oracle and sequential RL stages and report their range. We find a clear separation between the two training paradigms: SFT produces substantially larger parameter changes, typically $3$--$7\times$ those induced by RL and MMOPD, while mixed-SFT through \textsc{SDFT} yields the largest overall update. The same pattern appears in the parameter-space visualization in Figure~\ref{fig:pca_analysis}: the MMOPD checkpoint remains close to the final sequential model, whereas SDFT moves substantially farther away. These results suggest that RL makes relatively localized adjustments to the current policy, while SFT more extensively rewrites the model parameters.

SDFT also exhibits a consistent update direction across initializations. Applying the same oracle-mixture SFT objective from the base model and the final sequential checkpoint yields a cross-initialization update cosine of $0.77$, far above the $0.05$ similarity observed between unrelated SFT tasks. Together with its substantially larger update magnitude, this suggests that SDFT induces a strong, data-driven transformation that is largely independent of initialization. In contrast, RL and MMOPD produce much smaller updates that remain more closely tied to the starting policy.

\subsubsection{Complementary Roles of SFT and RL}

SFT provides the behavioral initialization needed for subsequent optimization by teaching the model the task interface, valid output format, and basic action space. As shown in Figure~\ref{fig:sft-fail}, directly applying RL to search tasks without prior SFT decreases the valid-format rate and increases truncation, indicating that the model lacks the basic interaction behavior required for effective RL training.

The main capability gains, however, come from RL-based optimization. RL builds on the SFT initialization to improve reasoning, decision making, and task-specific competence beyond behavior cloning. In Table~\ref{tab:main_mopd}, sequential training achieves strong performance across multiple domains, whereas SDFT produces uneven gains and substantial forgetting, for example, improving multi-hop search from 25.0 to 38.0 while reducing $\tau^3$-Telecom from 45.9 to 21.7 and IF-Eval from 84.8 to 53.6. MMOPD further strengthens several capabilities while better preserving performance across domains. These results suggest that SFT primarily establishes valid task behavior, whereas RL and MMOPD are responsible for advancing and consolidating task capabilities.
\begin{finding}[label=fin:sft-rl]{SFT establishes behavior; RL refines capability}
SFT induces large, directionally consistent updates that establish valid task behavior but can overwrite existing capabilities, whereas RL and MMOPD make smaller, policy-local updates that refine and consolidate task competence.
\end{finding}

\section{Ours - Mixture of Low-Rank Experts}

\noindent\textbf{Motivation.}
Guided by the analyses above, we distill three key properties that motivate our design.
(1) SDFT rapidly consolidates heterogeneous trajectories into a shared capability centroid, capturing common behaviors across domains.
(2) Different tasks induce partially misaligned optimization directions, making a single shared parameterization inherently prone to cross-stage interference.
(3) RL produces substantially smaller and more policy-local updates than SFT, suggesting that the remaining environment-specific specialization can be represented as lightweight residuals.
Together, these observations motivate a natural decomposition for ACL: transferable capabilities should be consolidated into a shared backbone, while interfering stage-specific residuals should be preserved separately.
Based on this principle, we propose Mixture of Low-Rank Experts (MLE).

\noindent\textbf{Architecture.}
MLE starts from an SDFT checkpoint trained over trajectories from multiple domains, which serves as the shared capability substrate.
We then freeze the backbone and instantiate a separate LoRA~\citep{hu2022lora} expert for each stage, optimized with RL in its corresponding environment.
Conceptually, each stage-specific policy is decomposed as
$\theta_k=\theta_{\mathrm{shared}}+\Delta\theta_k$,
where the small RL residual $\Delta\theta_k$ is parameterized by a low-rank adapter.
Crucially, these residuals are parameter-isolated: learning a new stage introduces a new lightweight LoRA expert without modifying the shared backbone or previously learned experts, thereby preventing subsequent stages from overwriting acquired specialization.

At inference time, MLE routes to the corresponding expert using the interaction environment available in the agent execution context, without requiring benchmark identities or ground-truth task labels.
Unlike shared-weight consolidation methods that seek a single compromise among conflicting task optima, MLE preserves incompatible specializations in separate low-rank subspaces while sharing the vast majority of model parameters.
Thus, parameter isolation provides continual retention, while low-rank adaptation makes this isolation inexpensive, yielding a unified deployable agent with only a small additive parameter cost per stage.

\noindent\textbf{Results.} As shown in Table~\ref{tab:main_mopd}, our method exhibits a clear advantage, particularly on the two more important agentic tasks. We can see that MMOPD performs well on E-commerce but only moderately on search, whereas SDFT performs well on search but only moderately on E-commerce; in contrast, our method approaches the performance of independently trained task experts across all in-domain benchmarks. 
More encouragingly, it also delivers strong out-of-domain performance, achieving the highest reported scores on GPQA, single-hop search, multi-hop search, and the mock domain among the evaluated methods. 
These results suggest that MLE’s benefits extend beyond retaining task-specific capabilities to supporting generalization beyond the training distributions. 

\section{Limitations \& Future Work
}

Our experiments mainly focus on \texttt{Qwen3-8B-Base} and a fixed four-stage curriculum. This controlled setting supports systematic comparisons, but the extent to which the observed forgetting patterns and method trade-offs generalize across model families and training orders remains unclear. In the future, extending \textsc{ACLArena} to additional agent environments, longer training sequences, and alternative curricula would help distinguish broadly recurring phenomena from effects specific to the present setup.

Besides, our proposed MLE uses observable environment context to select an expert. Its effectiveness when this context is ambiguous, changes during an interaction, or requires combining multiple specializations within a single episode remains untested. Promising directions include learned routing, dynamic expert composition, and adapter consolidation, evaluated jointly for capability retention, transfer, memory usage, and inference latency.

\section{Conclusion}

We introduce \textsc{ACLArena} to study forgetting and transfer across heterogeneous stages of post-training. Within this framework, we first examine sequential training and implement detailed diagnose. Based on existing observations, we systematically explore MMOPD, SDFT, and model merging as practical ACL strategies from different dimensions. Our analyzes reveal partially aligned task-specific updates, persistent capability trade-offs, and the complementary roles of SFT and RL. Guided by these findings, we propose MLE, which combines SDFT with LoRA-based RL experts to preserve task-specific adaptation and reduce interference.

\bibliographystyle{plainnat}
\bibliography{main}

\clearpage

\appendix
\appendix
\section*{Appendix}

\startcontents[appendix]
\begingroup
\titlecontents{lsection}[0em]
  {\addvspace{0.7em}\sffamily\bfseries\color{uclablue}}
  {\contentslabel{1.6em}}
  {\hspace*{-1.6em}}
  {\hspace{0.6em}\color{uclablue!35}\dotfill\color{uclablue}\contentspage}
\titlecontents{lsubsection}[2.0em]
  {\addvspace{0.15em}\small\sffamily\bfseries\color{uclablue}}
  {\contentslabel{2.6em}}
  {\hspace*{-2.6em}}
  {\hspace{0.6em}\color{uclablue!35}\dotfill\color{uclablue}\contentspage}
\printcontents[appendix]{l}{1}{\setcounter{tocdepth}{2}}
\endgroup
\vspace{1em}

\section{ACL Notations}
We study continual post-training across heterogeneous agent environments, where each stage introduces a distinct capability and interaction protocol. At deployment time, the environment context, including the system instruction and available tool schemas, is observable to the agent, while benchmark identities and ground-truth task labels are not provided. Our setting therefore focuses on preserving and composing capabilities across sequential post-training stages rather than inferring a latent task identity from an otherwise indistinguishable input distribution. Table \ref{tab:model_overview} demonstrates the settings of each paradigm in detail.

\begin{table*}[h]
\centering
\caption{%
\textbf{Overview of the models and capability-consolidation methods.}
Per-task experts are trained independently from the pretrained base model,
whereas sequential checkpoints inherit all preceding training stages.
MMOPD and SDFT are initialized from the final sequential checkpoint.
Model merging (MM) is training-free and directly averages the four
sequential training checkpoints.
MLE initializes from SDFT and trains a separate LoRA expert for each domain
using RL.
}
\label{tab:model_overview}
\setlength{\tabcolsep}{7pt}
\renewcommand{\arraystretch}{1.15}
\begin{tabularx}{\textwidth}{
    l
    l
    X
    l
}
\toprule
\textbf{Model}
& \textbf{Initialization}
& \textbf{Training source}
& \textbf{Training method} \\
\midrule

Base
& Pretrained
& --
& -- \\
\midrule

\rowcolor{black!10}
\multicolumn{4}{l}{\emph{Oracle Training}} \\

Expert$_{\text{Math}}$ (=Seq-${\text{Math}}$)
& Base
& DAPO-Math-17K 
& RL \\

Expert$_{\text{Search}}$
& Base
& Natural Question
& SFT $+$ RL \\

Expert$_{\text{E-commerce}}$
& Base
& $\tau^3$-Bench (retail) 
& SFT $+$ RL \\

Expert$_{\text{IF}}$
& Base
& Instruction Following 
& RL \\
\midrule

\rowcolor{black!10}
\multicolumn{4}{l}{\emph{Sequential Training}} \\

Seq-${\text{Math}}$ (=Expert$_{\text{Math}}$)
& Base
& DAPO-Math-17K  
& RL \\

Seq-${\text{SeaSFT}}$
& Seq-${\text{Math}}$
& Natural Question
& SFT \\

Seq-${\text{Search}}$
& Seq-${\text{SeaSFT}}$
& Natural Question 
& RL \\

Seq-${\text{E-comSFT}}$
& Seq-${\text{Search}}$
& $\tau^3$-Bench (retail)  
& SFT \\

Seq-${\text{E-commerce}}$
& Seq-${\text{E-comSFT}}$
& $\tau^3$-Bench (retail)  
& RL \\

Seq-${\text{IF}}$ (=Seq-${\text{Final}}$)
& Seq-${\text{E-commerce}}$
& IFTrain
& RL \\

\midrule

MMOPD
& Seq-${\text{Final}}$
& Mixed prompts with task-specific teachers
& OPD \\

\midrule

Base + SDFT
& Base
& Filtered trajectories generated by per-task expert models
& SFT \\

Seq + SDFT
& Seq-${\text{Final}}$
& Filtered trajectories generated by per-task expert models
& SFT \\

\midrule

MM
& \makecell[l]{N/A\\(training-free)}
& All four sequential-stage checkpoints
& Model Merging \\

\midrule

MLE
& SDFT
& Per-domain environments
& LoRA-RL \\

\bottomrule
\end{tabularx}
\end{table*}

\section{Training Details}

\subsection{Tool-calling Format}
Prior work often adopts heterogeneous tool-calling formats across agentic tasks: some datasets express function calls as free-form code snippets, while others use ad hoc JSON schemas or task-specific markup (\textit{e.g.,} \texttt{<search>}). This heterogeneity introduces two practical problems. First, the model must learn multiple surface syntaxes for what is semantically the same operation, which fragments the training signal and hinders cross-task transfer. Second, inconsistent formats complicate downstream parsing, making it difficult to build a single, reliable execution harness for evaluation and deployment.

To address these issues, we normalize the tool-call format of all tasks to the Qwen3 tool-call template. Under this convention, the available tools are declared in the system prompt as JSON schemas, and every invocation is emitted as a JSON object, containing the function name and an arguments dictionary, enclosed within dedicated \texttt{<tool\_call>} and \texttt{</tool\_call>} tags; execution results are returned to the model wrapped in corresponding \texttt{<tool\_response>} tags. During data construction, we convert every source dataset into this schema: function signatures are re-serialized into the unified JSON declaration format, and gold tool invocations are rewritten as structured JSON calls, with malformed or unparseable instances filtered out.

\subsection{Policy Optimization Algorithms}

All tasks in this work are optimized with reinforcement learning under the Group Sequence Policy Optimization (GSPO)~\citep{zheng2025group} algorithm, which we adopt as a single unified RL backbone across every stage of sequential training. Unlike token-level methods such as GRPO, whose per-token importance weights are high-variance and can accumulate into unstable gradients over long trajectories, GSPO fundamentally changes the unit of clipping by operating at the sequence level. It defines a single length-normalized importance ratio for the whole response,
\begin{equation}
s_i(\theta)
=
\exp\!\left(
\frac{1}{T_i}
\sum_{t=0}^{T_i-1}
\log w_{i,t}(\theta)
\right)
=
\left(
\frac{\pi_\theta(y_i \mid x)}
{\pi_{\theta_{\texttt{old}}}(y_i \mid x)}
\right)^{\!1/T_i},
\end{equation}
where $x$ denotes a single prompt and $\{y_i\}_{i=1}^{G}$ the group of $G$ responses that the rollout policy samples for that same prompt, so $i$ indexes responses within one group rather than distinct prompts. $T_i$ is the length of $y_i$ in tokens, and $\theta_{\texttt{old}}$ denotes the rollout policy parameters, \textit{i.e.,} the parameters held fixed while the group was sampled and against which the update is importance-weighted. The quantity $w_{i,t}(\theta)=\pi_\theta(y_{i,t}\mid x,y_{i,<t})/\pi_{\theta_{\texttt{old}}}(y_{i,t}\mid x,y_{i,<t})$ is the per-token importance ratio, which GSPO aggregates into the single sequence-level ratio $s_i(\theta)$ above rather than clipping token by token. The algorithm then applies clipping once per sequence,
\begin{equation}
\ell_i^{\text{GSPO}}(\theta)
=
\min\!\left(
s_i(\theta) A_i,\;
\mathrm{clip}\!\left(s_i(\theta), 1-\varepsilon, 1+\varepsilon\right) A_i
\right),
\end{equation}
where $A_i$ is the trajectory advantage of $y_i$, computed group-wise as the normalized deviation of its trajectory-level reward from the mean reward of the $G$ responses in its group, and $\varepsilon$ is the clipping threshold bounding how far $s_i(\theta)$ may move from $1$ before the update is truncated. With this sequence-level formulation, all tokens within a trajectory share the same clipped update. This aligns the unit of importance sampling with the unit of reward and enforces strong sequence-level coherence, thereby suppressing high-variance token outliers and yielding substantially more stable optimization in long-horizon agentic reinforcement learning. We therefore use GSPO with identical hyperparameters across all tasks unless otherwise noted.

\begin{figure*}[t]
\centering
\includegraphics[width=0.9\textwidth]{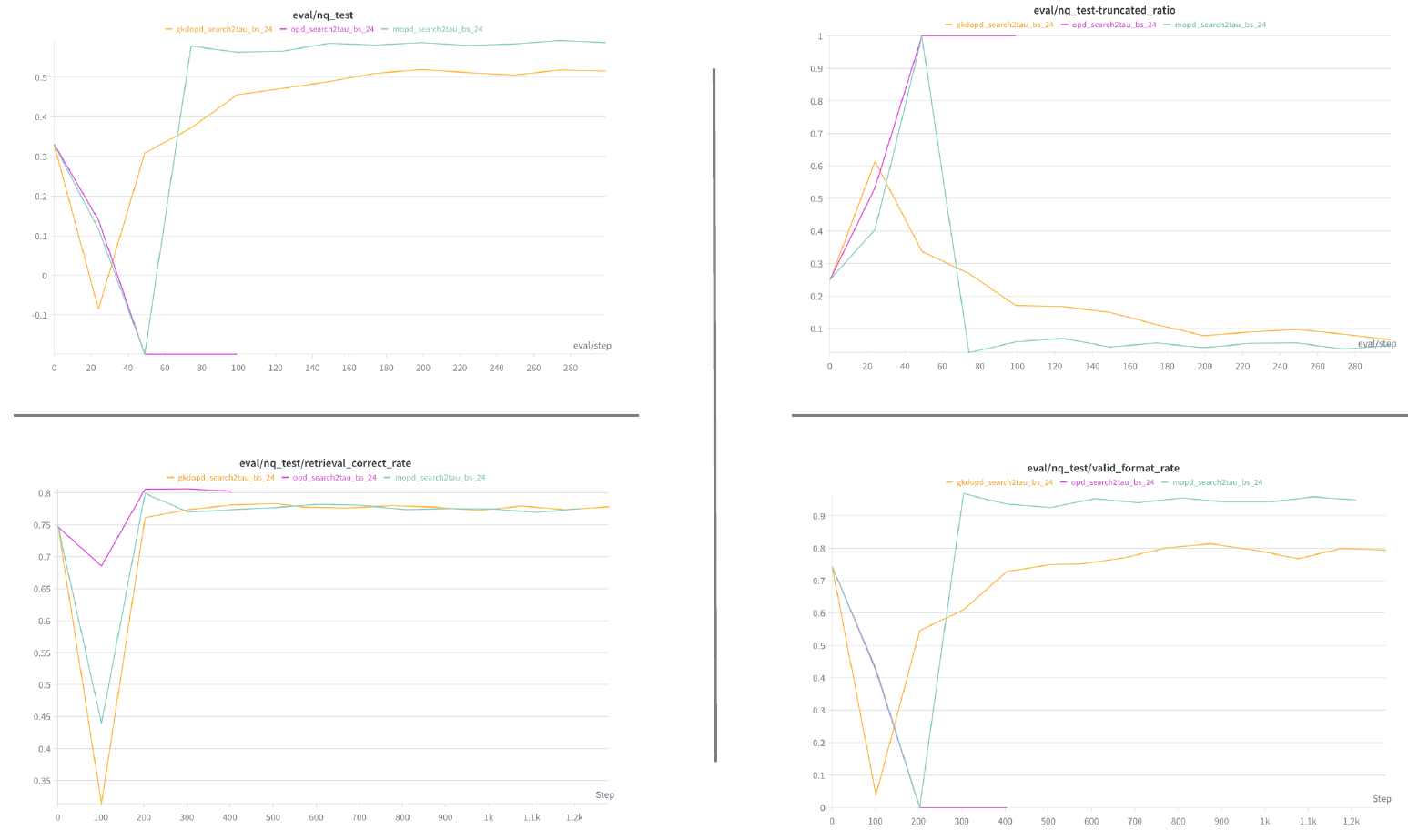}
\caption{Training collapse of vanilla OPD on the agentic search task. The green, purple, and yellow curves represent Mixed OPD, reverse-KL-based OPD, and forward-KL-based OPD, respectively. The green curve clearly demonstrates the stable training dynamics of Mixed OPD.}
\label{fig:opd-fail}
\end{figure*}

\begin{figure*}[t]
\begin{subfigure}[t]{0.49\linewidth}
    \centering
    \includegraphics[width=\linewidth]{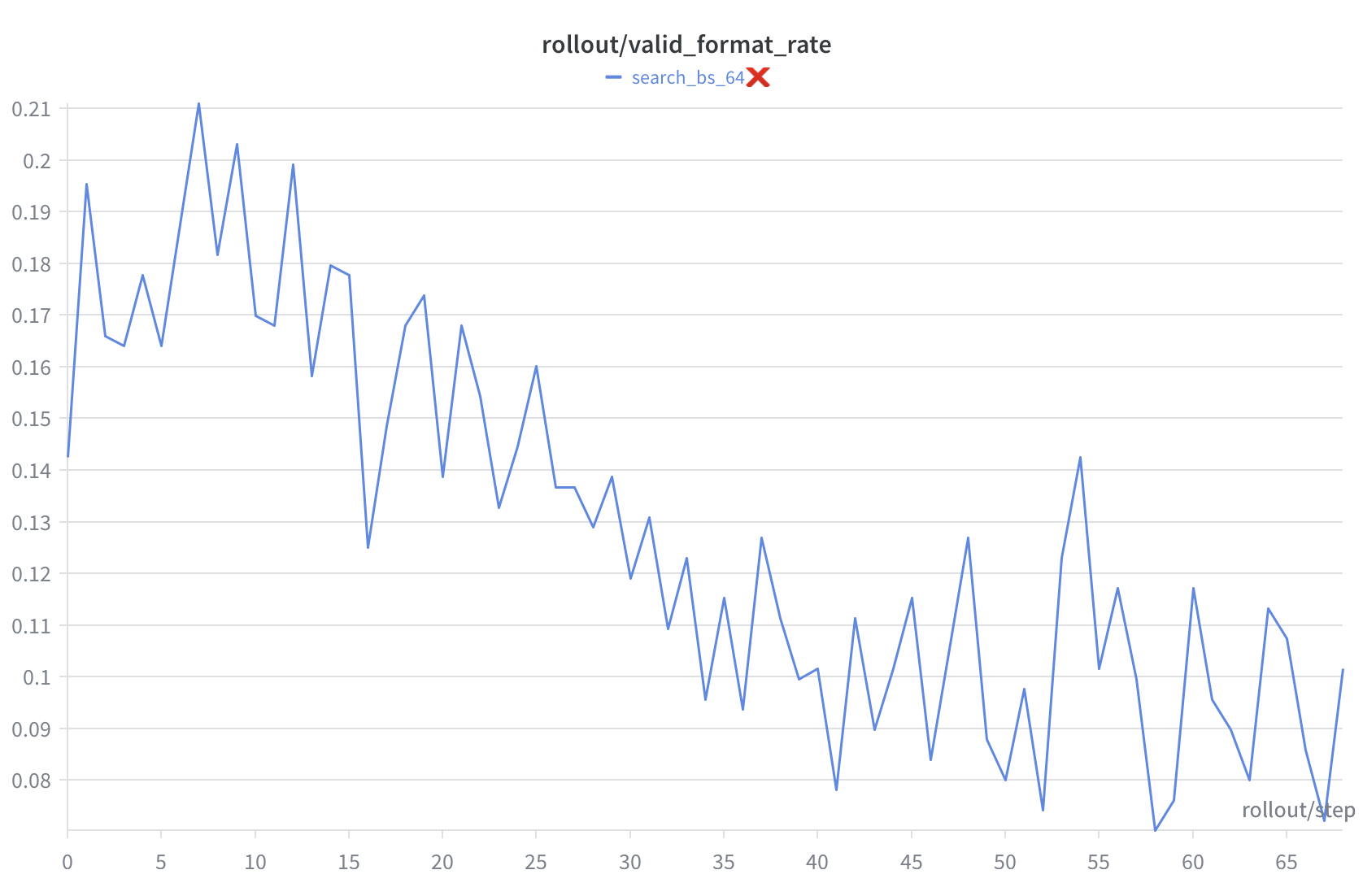}
    \caption{Format errors resulting from direct RL training on search tasks without prior SFT. }
\end{subfigure}
\hfill
\begin{subfigure}[t]{0.49\linewidth}
    \centering
    \includegraphics[width=\linewidth]{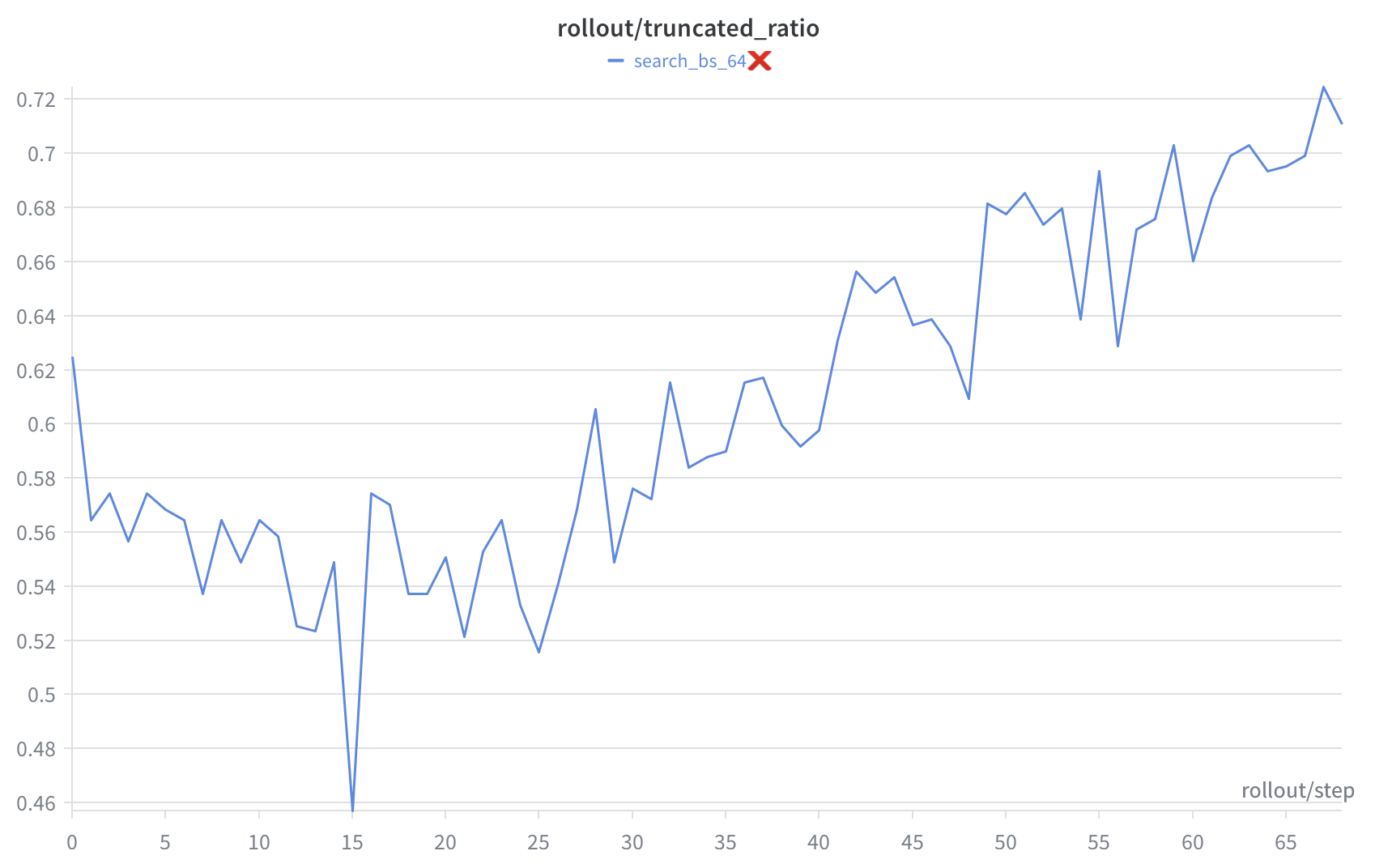}
    \caption{Increasing truncation ratios resulting from direct RL training on search tasks without prior SFT.}
\end{subfigure}
\caption{\textbf{Failure modes of direct RL without prior SFT on search tasks.} Direct RL leads to (a) frequent format violations and (b) increasingly truncated responses , highlighting the role of SFT in establishing valid task behavior before RL refinement.}
\label{fig:sft-fail}
\end{figure*}

\subsection{Task \& Environment Details}

\noindent\textbf{Math Reasoning.} It is trained by RL on DAPO-Math-17K~\citep{yu2026dapo}. Data is sourced from the web and official competition homepages through a combination of web scraping and manual annotation. Answers in math datasets typically appear in a variety of formats, such as expressions, formulas, and numbers, which makes it difficult to design comprehensive rules for parsing them. To provide accurate rule-based reward signals while minimizing the errors introduced by formula parsers, and inspired by AIME, we select and transform the answers into integers, which are easy to parse. After this selection and transformation, we obtain a dataset of 17K prompts, each paired with an integer answer.

\noindent\textbf{Agentic Search.}
We train the agent to invoke a search engine, thereby acquiring general question-answering capability that generalizes beyond the parametric knowledge stored in its weights. Rather than relying solely on memorized facts, the agent learns to formulate queries, retrieve relevant evidence, and ground its answers in the retrieved content, which is particularly important for questions requiring up-to-date or long-tail knowledge. For retrieval, we use the 2018 Wikipedia dump~\citep{karpukhin2020dense} as the knowledge source and E5~\citep{wang2022text} as the dense retriever. To ensure a fair comparison, we follow \citet{jin2025search} and set the number of retrieved passages to 3 across all retrieval-based methods, so that any performance differences can be attributed to the methods themselves rather than to varying amounts of retrieved context.

\noindent\textbf{Agentic E-commerce.}
We train the agent to interact with users and operate domain-specific tools under strict policy constraints, thereby acquiring the customer-service capability required for realistic e-commerce scenarios. It requires the agent to sustain a multi-turn dialogue with a user, invoke backend APIs to query and modify a persistent database, and adhere to domain-specific business rules throughout the interaction. For this task, we train on the retail domain of $\tau^3$-Bench~\citep{yao2024tau}, a benchmark that emulates dynamic conversations between an agent equipped with API tools and policy guidelines and a user simulated by a language model. Training uses the retail domain exclusively; the telecom and mock domains are never trained on and are held out for out-of-domain evaluation. Success is measured by comparing the database state at the end of the conversation against an annotated goal state, which provides a faithful and programmatically verifiable reward signal that reflects whether the agent has actually accomplished the user's request rather than merely producing plausible responses. To ensure a fair comparison, we adopt the same tool schemas, policy documents, and user-simulation configuration across all methods, so that performance differences reflect the agents' policy-following and tool-use abilities rather than discrepancies in the environment.

\noindent\textbf{Instruction Following.} We train the agent to follow precise, compositional instructions, thereby acquiring the controllability needed to satisfy explicit user-specified constraints. For this task, we build on IFTrain~\citep{pyatkin2025generalizing}, a collection of verifiable IF training constraints, each paired with a programmatic verification function that checks whether the constraint is satisfied. These executable verifiers make the task well suited to RL with verifiable rewards (RLVR): each output can be scored directly according to whether it satisfies the specified constraints, without relying on a learned reward model or an LLM-based judge.

\section{More details on MOPD}
\label{sec:supp_mopd}

\subsection{Failure of Vanilla OPD}

Figure \ref{fig:opd-fail} provides a comprehensive comparison of GKD-OPD (forward-KL only), OPD (reverse-KL only), and our proposed MOPD (mixed) on the task of distilling Search back into E-commerce.

First, we observe that OPD collapses rapidly: its truncate ratio climbs all the way to 100\%, its valid ratio falls to 0, and its final test-set performance drops to 0, even though retrieval correctness is still partially preserved. 
A closer analysis reveals that on content-related tokens the student and teacher are largely consistent, but on certain critical tokens, such as transitional or connective words and termination symbols, the teacher assigns very low probability, with the advantage on the order of $-$10. 
This clearly indicates that the teacher suppresses the student's critical outputs, preferring to continue reasoning and deliberating; as a result, the student fails to learn decisiveness in answer formatting, search syntax, and termination tokens. 
Second, we observe that pure forward-KL, while stable and free from collapse, has limited distillation capacity: it can cover the teacher's top-$q$ but cannot learn these critical decisions. 
Finally, our mixed OPD combines the strengths of both: it not only distills stably but also sharpens critical decisions.

\subsection{Implementation Details}
In our implementation, the coefficient $\gamma$ balancing the reverse-KL and forward-KL terms is set to $0.5$, the rollout batch size to $32$, the group number to $8$, and the global batch size to $256$. Each domain uses $3{,}000$ user prompts for OPD, with the OPD entropy threshold set to $0.3$ and top-$q$ set to $16$. As for training, we stop at 100 steps where the rewards of three tasks have converged already.

\section{More details on SDFT}

After obtaining the oracle model for each task, we first sample high-quality trajectories via rollout for training. Specifically, we sample 16 times for each user prompt in the dataset of each task, and then obtain high-quality trajectories through filtering. Finally, we obtain 156,285 trajectories for math, 157,750 for search, 19,573 for cs, and 225,681 for if. As for training, we stop at 1k steps where the loss has converged already.

\section{More details on Model Merging}

\begin{table*}[t]
\centering
\caption{%
\textbf{Model merging weight sweep.}
We merge five checkpoints into one model by linear weight averaging.
Rows correspond to different merge-weight allocations.
Balanced merging tests whether equal task contribution is sufficient, while
biased merging tests whether assigning a dominant weight to one checkpoint
improves average performance.
\textbf{Higher Avg indicates better overall performance across tasks.}}
\label{tab:model_merging_weights}
\setlength{\tabcolsep}{4pt}
\renewcommand{\arraystretch}{1.1}
\small
\begin{tabular}{l ccccc cccc c}
\toprule
\multirow{2}{*}{\textbf{Merge weights}}
& \multicolumn{5}{c}{\textbf{Weights}}
& \multicolumn{4}{c}{\textbf{ID evaluation} ($\uparrow$)}
& \multirow{2}{*}{\makecell{Avg\\($\uparrow$)}} \\
\cmidrule(lr){2-6}
\cmidrule(lr){7-10}
& Base & Math & Search & E-commerce & IF
& \makecell{AIME26\\(avg@16)}
& Search
& $\tau^3$-Retail
& IF
& \\
\midrule

Uniform
& 0.2 & 0.2 & 0.2 & 0.2 & 0.2
& 10.62 & 37.0 & 11.11 & 67.8
& 31.63 \\

Base-heavy
& 0.6 & 0.1 & 0.1 & 0.1 & 0.1
& 6.25 & 22.0 & 5.84 & 53.4
& 21.87 \\

Math-heavy
& 0.1 & 0.6 & 0.1 & 0.1 & 0.1
& 18.12 & 18.5 & 1.97 & 49.7
& 22.07 \\

Search-heavy
& 0.1 & 0.1 & 0.6 & 0.1 & 0.1
& 10.83 & 44.5 & 0.89 & 61.9
& 29.53 \\

E-commerce-heavy
& 0.1 & 0.1 & 0.1 & 0.6 & 0.1
& 8.54 & 16.3 & 17.56 & 40.3
& 20.68 \\

IF-heavy
& 0.1 & 0.1 & 0.1 & 0.1 & 0.6
& 0.83 & 29.1 & 5.90 & 79.5
& 28.83 \\

Task-balanced, no Base/SFT
& 0.0 & 0.25 & 0.25 & 0.25 & 0.25
& 12.92 & 10.3 & 3.57 & 69.7
& 24.12 \\

\bottomrule
\end{tabular}
\end{table*}

Table~\ref{tab:model_merging_weights} highlights the asymmetry and interference inherent in linear model merging. Uniform merging achieves the highest average performance at 31.63, followed by Search-heavy merging at 29.53, while emphasizing a particular checkpoint generally improves its corresponding capability at the expense of others. For example, Search-heavy merging raises search performance to 44.5, and E-commerce-heavy merging improves e-commerce performance to 17.56, but both exhibit clear degradation on other tasks. Similarly, IF-heavy merging achieves the strongest IF performance at 79.5 while remaining weaker on several other capabilities. Overall, the results suggest that adjusting merge weights primarily shifts the trade-off among capabilities rather than eliminating interference: no single linear weighting scheme consistently preserves strong performance across all tasks.

\section{Evaluation Details}
\label{appendix_eval_detial}

All models are evaluated under a consistent inference setup. We serve models with SGLang~\citep{zheng2024sglangefficientexecutionstructured}, using \texttt{mem-fraction-static=0.85} and the Qwen tool-call parser. Unless otherwise specified, we use greedy decoding with temperature 0. For MMLU-Redux, we use chain-of-thought prompting with temperature 0.7 and a maximum generation length of 16,384 tokens. GPQA-Diamond uses chain-of-thought prompting with temperature 0 and a maximum generation length of 30,000 tokens. For AIME 2026, we sample 16 responses per problem with temperature 1.0 and top-$p$ 0.7, using a maximum generation length of 30,000 tokens, and report avg@16. IF-Eval and IF-Bench also use a maximum generation length of 30,000 tokens.
For search, the agent is allowed up to five interaction turns and 2,048 newly generated tokens per turn. Retrieval is performed using an E5 retriever~\citep{wang2024textembeddingsweaklysupervisedcontrastive} and a FAISS index~\citep{douze2024faiss} over Wikipedia-18~\citep{karpukhin2020densepassageretrievalopendomain}, with the top three passages returned for each query.
For $\tau^3$-Bench, we use the basic split, a maximum of 200 interaction steps, a maximum generation length of 8,192 tokens, temperature 0, and \texttt{<|im\_end|>} as the stop sequence. User simulation uses GLM-4.7-Flash~\citep{5team2025glm45agenticreasoningcoding}.

\section{More Details on Model-level Analysis}
\label{app:model-level}

\begin{figure}[t]
\centering
\includegraphics[width=0.48\textwidth]{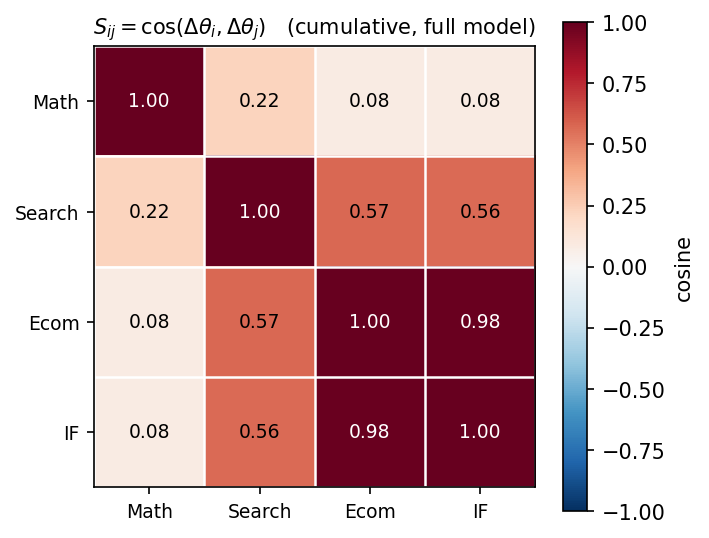}
\caption{Cosine similarity between domain task vectors $\Delta\theta_k=\theta_k-\theta_0$.}
\label{fig:cos}
\end{figure}

\subsection{Stronger Evidence}
To better validate Finding 1, Figure~\ref{fig:cos} reports the pairwise cosine similarity between the four domains' task vectors in weight space. Following the task-arithmetic definition, each task vector is the displacement of a domain's fine-tuned checkpoint from the shared base model, $\Delta\theta_k=\theta_k-\theta_0$ (with $\theta_0=$ \texttt{Qwen3-8B-Base}), and the matrix entries are $S_{ij}=\langle\Delta\theta_i,\Delta\theta_j\rangle/(\lVert\Delta\theta_i\rVert\,\lVert\Delta\theta_j\rVert)$. The four columns/rows are the domain RL endpoints along our sequential training chain.

For every parameter tensor we load the base and the four endpoint checkpoints, form the per-tensor difference $\Delta\theta_k$, and accumulate the full-model Gram matrix $\langle\Delta\theta_i,\Delta\theta_j\rangle$ by streaming tensor-by-tensor across the whole network. The matrix is then normalized to unit-norm cosines and rendered with a diverging color scale centered at 0 (white $=$ orthogonal, red $=$ aligned).

The structure is dominated by the cumulative, sequential nature of training. Math is nearly orthogonal to the later domains ($\cos\approx0.08$ to both E-commerce and IF): the earliest task direction is almost entirely rotated away by subsequent stages. E-commerce and IF are near-collinear ($\cos=0.98$), because IF is trained on top of the service checkpoint and adds only a small further displacement, so its total vector from base barely departs from where E-commerce already sits. Search sits in between ($\cos\approx0.57$ with the two later domains, $0.22$ with Math).

Because the domains are trained sequentially, these task vectors are nested (each later vector contains the earlier ones as a prefix), so the high off-diagonal values reflect the accumulation geometry of the training trajectory rather than intrinsic task similarity, the large cosines should not be read as ``these tasks are similar." When the shared prefix is removed and each domain's own RL increment is isolated, the four directions become mutually near-orthogonal ($\lvert\cos\rvert\le0.004$; companion panel), indicating that each domain carves out an essentially independent direction in weight space.

\subsection{Beyond a Particular Curriculum}
The observed capability conflict is not solely induced by our chosen
training order. All oracle experts are independently optimized from the
same base model and therefore contain no sequential-training history.
Nevertheless, their cross-domain performance exhibits heterogeneous
transfer: for example, the Search expert substantially improves Search
while reducing $\tau^3$-Telecom, whereas the
E-commerce expert produces a different transfer profile on out-of-domain
search benchmarks. This observation is consistent with Figure~6, where
isolated task-specific updates are nearly orthogonal.
Together, these results provide order-independent evidence that
heterogeneous post-training tasks induce intrinsically distinct and
potentially conflicting optimization effects; the sequential curriculum
determines how such interference manifests, rather than being its sole
cause.

\section{More Details on Token-level Analysis}
\label{app:token-level}

\begin{figure*}[t]
\centering
\includegraphics[width=\textwidth]{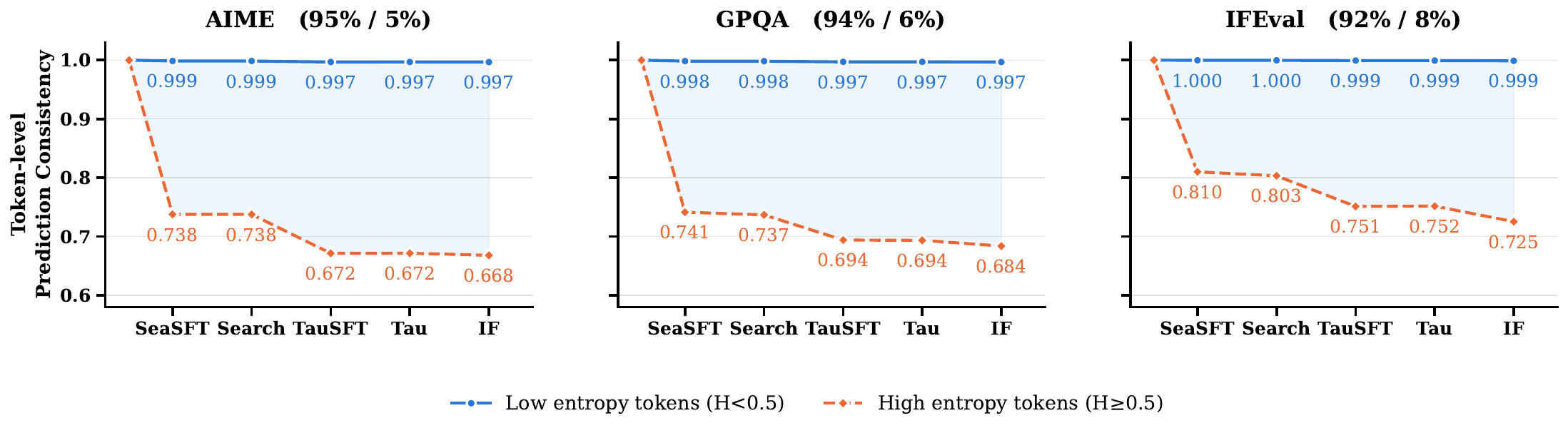}
\caption{Token-level prediction consistency along the training chain, split at
$H=0.5$, for all three benchmarks. The leftmost point is the math checkpoint itself
and is $1.0$ by definition; panel headings give the low- / high-entropy share of
positions.}
\label{fig:entropy-all}
\end{figure*}

\subsection{Measurement Protocol}

All checkpoints are compared on a single fixed set of token ids. We decode each
benchmark once with the math checkpoint using greedy decoding with temperature 0 and store the
generated token ids. Every subsequent checkpoint is then run
in a forward-only pass over the identical id sequence, so that at every position all
six checkpoints are conditioned on exactly the same prefix. This yields, per position,
each checkpoint's full next-token distribution over the same context.

Write $x=(x_1,\dots,x_T)$ for the stored ids and $p^c_t(\cdot)$ for the next-token
distribution that checkpoint $c$ assigns at position $t$ given the prefix $x_{<t}$. The \emph{entropy} of a position is
that of the math checkpoint $m$, computed exactly over the full vocabulary
($|V|=151{,}936$) by reducing the logits directly,
\begin{equation*}
H_t \;=\; -\!\!\sum_{v \in V} p^m_t(v)\,\log p^m_t(v),
\end{equation*}
so a position is labelled once, by the checkpoint that produced the trajectory, and
that label does not change with the checkpoint being compared against it. A position
is then \emph{consistent} for checkpoint $c$ when
$\arg\max_v p^c_t(v) = \arg\max_v p^m_t(v)$, and the consistency we report for a group
of positions is the fraction of them that are consistent.

\subsection{Splitting Positions}

Positions are split at $H = 0.5$ nats, which Table~\ref{tab:entropy-bins} shows is
where the behaviour changes: consistency stays above $0.97$ throughout the bins below
$0.2$, is still around $0.92$ in $[0.2,0.5)$, and only then falls steeply, to roughly
$0.76$, $0.61$ and $0.46$ in the successive bins above it. The degradation is smooth
in entropy, so no threshold is privileged; $0.5$ places the boundary at the knee.

Rates are token-weighted throughout: the fraction of positions in a group whose top-1
prediction is unchanged. The two groups are strongly unbalanced on every benchmark,
with low-entropy positions accounting for $95.2\%$, $93.6\%$ and $91.9\%$ of all
positions on AIME, GPQA and IF-Eval respectively.

Figure~\ref{fig:entropy-all} gives the resulting curves for all three benchmarks; the
main text reports GPQA. The two groups behave the same way on each of them, the
low-entropy majority is preserved almost exactly across the whole sequence of training
stages, while the high-entropy minority degrades monotonically, so the pattern is not
specific to the benchmark shown in the main text.

\begin{table*}[t]
\centering\small
\caption{Token-level prediction consistency by entropy bin. Each entry is the fraction of
positions in that bin at which the checkpoint still ranks the math checkpoint's token first.
The two shaded rows per benchmark are the aggregates plotted in Figure~\ref{fig:entropy-all};
they are token-weighted over the bins above them, not averages of the per-bin rates, note how
few positions the highest bins hold. Bins marked $\dagger$ have fewer than $500$ positions
and are not reliable.}
\label{tab:entropy-bins}
\begin{tabularx}{\textwidth}{@{}llrr*{5}{>{\centering\arraybackslash}X}@{}}
\toprule
Benchmark & Entropy bin & $n$ & \% of tokens & Seq-${\text{SeaSFT}}$ & Seq-${\text{Search}}$ & Seq-${\text{E-comSFT}}$ & Seq-${\text{E-commerce}}$ & Seq-${\text{IF}}$ \\
\midrule
\multicolumn{9}{l}{\textit{AIME} \ (348,828 positions)} \\
 & $[0,0.01)$ & 298,125 & 85.5\% & 0.9998 & 0.9998 & 0.9997 & 0.9997 & 0.9997 \\
 & $[0.01,0.05)$ & 16,947 & 4.9\% & 0.9980 & 0.9980 & 0.9943 & 0.9945 & 0.9941 \\
 & $[0.05,0.2)$ & 9,198 & 2.6\% & 0.9935 & 0.9927 & 0.9776 & 0.9776 & 0.9764 \\
 & $[0.2,0.5)$ & 7,720 & 2.2\% & 0.9600 & 0.9567 & 0.9118 & 0.9130 & 0.9088 \\
\rowcolor{gray!12}  & \textbf{$H<0.5$} & 331,990 & 95.2\% & 0.9986 & 0.9985 & 0.9968 & 0.9968 & 0.9967 \\
 & $[0.5,1)$ & 10,380 & 3.0\% & 0.8001 & 0.8003 & 0.7356 & 0.7341 & 0.7294 \\
 & $[1,2)$ & 6,153 & 1.8\% & 0.6425 & 0.6434 & 0.5739 & 0.5765 & 0.5750 \\
 & $[2,3)$$^\dagger$ & 303 & 0.09\% & 0.5314 & 0.5050 & 0.4554 & 0.4620 & 0.4587 \\
 & $[3,\infty)$$^\dagger$ & 2 & 0.00\% & 1.0000 & 1.0000 & 1.0000 & 1.0000 & 1.0000 \\
\rowcolor{gray!12}  & \textbf{$H\geq0.5$} & 16,838 & 4.8\% & 0.7377 & 0.7377 & 0.6715 & 0.6716 & 0.6681 \\
\midrule
\multicolumn{9}{l}{\textit{GPQA} \ (1,229,677 positions)} \\
 & $[0,0.01)$ & 1,030,032 & 83.8\% & 1.0000 & 1.0000 & 0.9999 & 0.9999 & 0.9999 \\
 & $[0.01,0.05)$ & 53,827 & 4.4\% & 0.9971 & 0.9972 & 0.9938 & 0.9940 & 0.9944 \\
 & $[0.05,0.2)$ & 36,340 & 3.0\% & 0.9861 & 0.9856 & 0.9771 & 0.9771 & 0.9748 \\
 & $[0.2,0.5)$ & 31,166 & 2.5\% & 0.9601 & 0.9567 & 0.9298 & 0.9294 & 0.9227 \\
\rowcolor{gray!12}  & \textbf{$H<0.5$} & 1,151,365 & 93.6\% & 0.9983 & 0.9982 & 0.9970 & 0.9970 & 0.9968 \\
 & $[0.5,1)$ & 42,773 & 3.5\% & 0.8116 & 0.8066 & 0.7700 & 0.7702 & 0.7592 \\
 & $[1,2)$ & 32,195 & 2.6\% & 0.6684 & 0.6641 & 0.6169 & 0.6158 & 0.6068 \\
 & $[2,3)$ & 3,256 & 0.26\% & 0.5494 & 0.5488 & 0.4653 & 0.4616 & 0.4619 \\
 & $[3,\infty)$$^\dagger$ & 88 & 0.01\% & 0.4773 & 0.4318 & 0.4886 & 0.4773 & 0.4432 \\
\rowcolor{gray!12}  & \textbf{$H\geq0.5$} & 78,312 & 6.4\% & 0.7414 & 0.7368 & 0.6941 & 0.6936 & 0.6838 \\
\midrule
\multicolumn{9}{l}{\textit{IFEval} \ (1,377,180 positions)} \\
 & $[0,0.01)$ & 1,125,401 & 81.7\% & 1.0000 & 1.0000 & 1.0000 & 1.0000 & 1.0000 \\
 & $[0.01,0.05)$ & 66,560 & 4.8\% & 0.9999 & 0.9999 & 0.9993 & 0.9993 & 0.9985 \\
 & $[0.05,0.2)$ & 38,522 & 2.8\% & 0.9987 & 0.9986 & 0.9959 & 0.9958 & 0.9926 \\
 & $[0.2,0.5)$ & 34,698 & 2.5\% & 0.9912 & 0.9900 & 0.9795 & 0.9798 & 0.9707 \\
\rowcolor{gray!12}  & \textbf{$H<0.5$} & 1,265,181 & 91.9\% & 0.9997 & 0.9997 & 0.9993 & 0.9993 & 0.9989 \\
 & $[0.5,1)$ & 45,363 & 3.3\% & 0.8958 & 0.8810 & 0.8614 & 0.8615 & 0.8438 \\
 & $[1,2)$ & 46,238 & 3.4\% & 0.7842 & 0.7825 & 0.7191 & 0.7199 & 0.6913 \\
 & $[2,3)$ & 15,547 & 1.1\% & 0.6889 & 0.6871 & 0.5947 & 0.5962 & 0.5580 \\
 & $[3,\infty)$ & 4,851 & 0.35\% & 0.6425 & 0.6471 & 0.5283 & 0.5271 & 0.4785 \\
\rowcolor{gray!12}  & \textbf{$H\geq0.5$} & 111,999 & 8.1\% & 0.8100 & 0.8033 & 0.7512 & 0.7517 & 0.7254 \\
\bottomrule
\end{tabularx}
\end{table*}

\subsection{Reading the Per-bin Table}

Table~\ref{tab:entropy-bins} is monotone in two directions on all three benchmarks:
within a column, consistency falls as entropy rises; within a row, it falls along the
training order, with $\text{\textsc{Seq-SeaSFT}} \approx \text{\textsc{Seq-Search}} > \text{\textsc{Seq-E-comSFT}} \approx
\text{\textsc{Seq-E-commerce}} \gtrsim \text{\textsc{Seq-IF}}$. Both hold at every reliable cell.

\end{document}